\documentclass{article} 
\usepackage[preprint]{iclr2027_conference}
\usepackage{times}

\usepackage{amsmath,amsfonts,bm}

\def\eqref#1{equation~\ref{#1}}

\def\1{\bm{1}}

\DeclareMathAlphabet{\mathsfit}{\encodingdefault}{\sfdefault}{m}{sl}
\SetMathAlphabet{\mathsfit}{bold}{\encodingdefault}{\sfdefault}{bx}{n}

\usepackage{hyperref}       
\usepackage{xurl}            
\usepackage{booktabs}       
\usepackage{amsfonts}       
\usepackage{nicefrac}       
\usepackage{microtype}      
\usepackage{xcolor}         

\usepackage{algorithm}
\usepackage{algorithmic}
\usepackage{amsmath}
\usepackage{multirow}
\usepackage{makecell}
\usepackage{enumitem}
\usepackage{subcaption}
\usepackage{graphicx}
\usepackage{placeins} 
\usepackage{tabularx}
\usepackage{pifont}
\usepackage{wasysym}

\title{NeuroAct: Advancing EEG Analysis and Decoding via Dual-Loop Autonomous Research}

\author{Wei Xiong$^{1,2}$, Jiangtong Li$^1$, Han Zhao$^3$, Linxiao Yang$^2$, Jie Li$^{4,1}$, Changjun Jiang$^1$ \\
\textnormal{$^1$Tongji University \quad $^2$Ant Group \quad $^3$Shanghai Jiao Tong University}\\
\textnormal{$^4$Translational Research Center,}
\textnormal{Shanghai Yangzhi Rehabilitation Hospital (Shanghai}\\
\textnormal{\space Sunshine Rehabilitation Center), Tongji University}
}

\begin{document}
\raggedbottom

\maketitle

\begin{abstract}
EEG analysis and decoding underpin brain-computer interfaces, cognitive assessment, and sleep monitoring. 
Fixed pipelines and reward-driven search do not explicitly determine how signal structure, representations, predictor behavior, and training dynamics contribute to performance or generalization gaps, leaving intervention choice underdetermined. 
We introduce NeuroAct, an autonomous research agent framework for EEG analysis and decoding. 
NeuroAct implements \textbf{dual-loop autonomous research} across EEG foundation models, task-specific predictors, and model combinations. 
In an inner loop, research, development, and review roles use evidence on signal structure, representations, predictor behavior, and training dynamics to test hypotheses, select analyses or interventions, and refine candidates using performance, stability, and cost feedback. 
An outer loop distills outcomes into conditional, testable recommendations with task and model conditions, then validates and revises them on subsequent tasks.
We evaluate four settings: (A) EEG General Analysis, (B) Cross-task Model Optimization, (C) Single-task Open Research, and (D) Collaborative Prediction.
In Setting A, NeuroAct's 80.88 macro-average exceeds BrainAgent's 63.13 by 17.75 points.
In Setting B, NeuroAct raises EEGPT's mean balanced accuracy~(BA) by 7.34 points and HBN Pearson's \(r\) from 0.6894 to 0.7362; in Setting C, it improves ADFTD BA by 16.33 points; and in Setting D, it raises Workload BA from 79.69\% to 85.02\%.
Experience reuse raises BCIC-IV-2a BA from 34.66\% to 50.81\% while reducing LLM tokens by 64.02\%, indicating recommendations improve cross-task performance.
NeuroAct enables evidence-guided EEG research improved by cross-task experience.
\end{abstract}




%


\section{Introduction}
\label{sec:introduction}

EEG analysis and decoding support cognitive-state assessment, sleep monitoring, and brain-computer interfaces. 
Pretrained EEG foundation models~(EEG-FMs) provide reusable representations across diverse tasks~\citep{jiang2024labram,wang2024eegpt,wang2025cbramod,elouahidi2025reve,doner2025luna}. 
Effective task adaptation requires assessing how signal structure, representations, predictor behavior, and training dynamics contribute to performance or generalization gaps, then selecting evidence that distinguishes competing explanations. 
Fixed pipelines and reward-driven optimization do not make this diagnostic decision explicit, motivating an \textit{\textbf{autonomous research agent that improves within-task EEG decoding and refines research strategies across tasks.}}

Task-specific EEG analysis must connect signal structure to how a model represents, predicts, and learns. 
Benchmark studies show model rankings and transfer depend on task and readout design, encoded features can differ from those used for prediction, and token preservation affects frozen-encoder decoding~\citep{tang2026eeg_feature_usage,sirca2026beyond_accuracy}. 
When refinement of a single EEG model yields diminishing gains, this diagnosis can test whether other models correct its errors on samples.
Error overlap determines whether complementary prediction merits a model-combination experiment.

Autonomous research systems combine hypothesis generation, experimentation, and interpretation in continuous workflows~\citep{lu2026ai_scientist}.
Model-development systems further use experimental feedback or accumulated experience to revise their approaches~\citep{alphaxiv2026openresearch}. 
In the EEG domain, BrainAgent orchestrates analysis tools~\citep{zhou2026brainagent}, BrainPilot combines domain knowledge and research skills~\citep{li2026brainpilot}, and NeuroWeaver searches task pipelines~\citep{wang2026neuroweaver}. 
For EEG research, open questions concern how diagnostics guide interventions across foundation-model adaptation, task-specific modeling, joint training, and model combination, and how cross-task validation and revision make conditional recommendations useful on subsequent tasks.

We introduce \textbf{NeuroAct}, an autonomous research agent framework for EEG analysis and decoding. 
NeuroAct implements a \textbf{dual-loop autonomous research} framework: an inner loop selects analyses and interventions from evidence on signal structure, representations, predictor behavior, and training dynamics, while an outer loop distills, validates, and revises conditional recommendations or experiences across research tasks.
Within the inner loop, the Coordinator directs the Researcher, Developer, and Reviewer to test competing diagnostic hypotheses through executable analyses or model changes, with predictive performance, stability, and cost informing the next action for each task. 
Each experience item is a conditional recommendation that records applicable EEG task and model conditions, a proposed action, supporting and counterevidence, and reassessment conditions. 
Subsequent tasks test its utility, while counterevidence narrows, suspends, or replaces the recommendation. 
Tab.~\ref{tab:agent_framework_comparison} positions this scope relative to existing agent frameworks.

\newcommand{\agentsupport}{\ding{51}}
\newcommand{\agentpartial}{\LEFTcircle}
\newcommand{\agentnosupport}{\ding{55}}
\newcommand{\agentunknown}{\textemdash}

\begin{table}[t]
    \caption{EEG analysis, decoding, and research guidance across agent frameworks~\citep{wang2024codeact,zhou2026brainagent,li2026brainpilot,wang2026neuroweaver}. \ding{51}: supported; \LEFTcircle: partial; ---: unspecified.}
    \label{tab:agent_framework_comparison}
    \centering
    \small
    \setlength{\tabcolsep}{4pt}
    \begin{tabularx}{\textwidth}{@{}Xccccc@{}}
        \toprule
        Scope / mechanism & CodeAct & BrainAgent & BrainPilot & NeuroWeaver & \textbf{NeuroAct} \\
        \midrule
        Primary scope
        & \makecell{General\\coding}
        & \makecell{Signal\\analysis}
        & \makecell{Brain\\research}
        & \makecell{Pipeline\\search}
        & \makecell{Analysis \&\\decoding} \\
        \midrule
        \multicolumn{6}{@{}l}{\textit{Research actions}} \\
        Code generation \& execution
        & \ding{51} & --- & \ding{51} & \ding{51} & \ding{51} \\
        EEG analysis workflows
        & --- & \ding{51} & \ding{51} & \LEFTcircle & \ding{51} \\
        Model editing \& training
        & \ding{51} & --- & \ding{51} & \ding{51} & \ding{51} \\
        EEG-FM adaptation \& fine-tuning
        & --- & --- & --- & --- & \ding{51} \\
        Joint multi-task training
        & --- & --- & --- & --- & \ding{51} \\
        Model combination
        & --- & --- & --- & --- & \ding{51} \\
        \midrule
        \multicolumn{6}{@{}l}{\textit{Feedback and experience}} \\
        Representation \& gradient analysis
        & --- & --- & \ding{51} & --- & \ding{51} \\
        Feedback-based iteration
        & \ding{51} & \LEFTcircle & \ding{51} & \ding{51} & \ding{51} \\
        Progressive capability disclosure
        & --- & --- & \ding{51} & --- & \ding{51} \\
        Cross-task experience revision
        & --- & --- & --- & --- & \ding{51} \\
        \bottomrule
    \end{tabularx}
\end{table}

We organize NeuroAct's evaluation around \textbf{four research settings}: \textit{(A) EEG General Analysis}, \textit{(B) Cross-task Model Optimization}, \textit{(C) Single-task Open Research}, and \textit{(D) Collaborative Prediction}.
The evaluation covers \textbf{six datasets} (HBN~\citep{alexander2017hbn,shirazi2024hbn_eeg}, Workload~\citep{zyma2019eegmat}, PhysioMI~\citep{schalk2004bci2000}, BCIC-IV-2a~\citep{tangermann2012bci_competition}, ADFTD~\citep{miltiadous2023adftd}, and HMC~\citep{alvarez2021hmc}), \textbf{three LLM backend families} (DeepSeek-V4 Flash~\citep{deepseek2026v4}, GPT-5.6 Luna~\citep{openai2026gpt56}, and Qwen3.8 Flash Next~\citep{qwen2026next}), and the available external baselines CodeAct~\citep{wang2024codeact}, BrainPilot~\citep{li2026brainpilot}, and BrainAgent~\citep{zhou2026brainagent}.
With DeepSeek-V4 Flash, NeuroAct reaches an 80.88 macro-average score in Setting A and exceeds 75 with Luna and Qwen on BrainBench~\citep{zhou2026brainbench}.
In Setting B, NeuroAct raises mean balanced accuracy~(BA) from 47.50 to 54.84 and HBN Pearson's \(r\) from 0.6894 to 0.7362.
NeuroAct gains 16.33 BA points on ADFTD in Setting C and raises Workload BA from the best single model's 79.69\% to 85.02\% in Setting D.
Experience reuse transfers a PhysioMI-derived recommendation to BCIC-IV-2a, raising test BA from 34.66\% to 50.81\% while reducing LLM tokens by 64.02\%.
Our contributions are:
\begin{itemize}[leftmargin=0.7em,rightmargin=0pt,labelsep=0.4em]
    \setlength{\itemsep}{1pt}
    \setlength{\parsep}{0pt}
    \item \textbf{Evidence-guided inner-loop EEG research.} NeuroAct tests alternatives and selects analyses or interventions using signal, representation, predictor, and training evidence.
    \item \textbf{Outer-loop conditional strategy improvement.} NeuroAct tests conditional recommendations across tasks and revises them using matched performance and cost comparisons when warranted.
    \item \textbf{Executable EEG analysis and decoding.} NeuroAct operationalizes evidence through EEG General Analysis, Cross-task Model Optimization, Single-task Open Research, and Collaborative Prediction.
\end{itemize}


\section{Related Work}
\label{sec:related_work}

\subsection{EEG Decoding Models}
Task-specific architectures including EEGNet and EEG Conformer encode local and temporal EEG structure~\citep{lawhern2018eegnet,song2023eegconformer}. 
Pretrained models including BENDR and EEGPT learn reusable representations, while other EEG-FMs differ in treating channels, signal structure, and recording heterogeneity~\citep{kostas2021bendr,yang2023biot,jiang2024labram,wang2024eegpt,wang2025cbramod,elouahidi2025reve,doner2025luna}. 
These approaches expand candidate space from specialized decoders to transferable EEG representations.

Benchmark studies show model rankings, transfer, and generalization vary with task, dataset, and readout design~\citep{banville2026neuralbench,xiong2026eegfmbench,liu2026eegfmcompass}. 
Adaptation studies identify cohort shift, data availability, channel adaptation, and training regime as consequential~\citep{nguyen2026stroke_adaptation,yang2026eegfm_evaluation,kokate2026channel_adaptation,kommineni2026generalization}. 
NeuroAct turns evidence on signal structure, representations, predictor behavior, and training dynamics, including cross-model error overlap, into diagnostic decisions selecting analyses or interventions across adaptation, shared-model training, and model combination rather than a fixed recipe.

\subsection{Autonomous Research Agents}
Autonomous research systems coordinate hypothesis generation, experimentation, and interpretation in sustained workflows~\citep{swanson2025virtual_lab,gottweis2026co_scientist,ghareeb2026robin}. 
Computational agents extend this workflow through executable tool use, candidate search, targeted refinement, repository-grounded development, diagnostic code updates, and training feedback, using execution outcomes to focus candidates~\citep{jiang2025aide,yamada2025aiscientistv2,nam2025mlestar,yang2025rdagent,li2026autosota,zhang2026reasoning_gradient,li2026ftdojo}. 
Long-horizon systems support asynchronous experiments and interactive debugging~\citep{hambardzumyan2026aira2} and allocate effort across competing plans using feedback and remaining resources~\citep{yu2026primescientist}.

Within EEG and brain-signal research, EEGAgent and BrainAgent orchestrate signal-processing tools and domain agents, while CogEEGAgent emphasizes deterministic cognitive-analysis execution and independent confirmation~\citep{zhao2026eegagent,zhou2026brainagent,hou2026cogeegagent}. 
BrainPilot combines knowledge, skills, and specialists for architecture development under fixed preprocessing and training, whereas NeuroWeaver searches domain-constrained pipelines through tree selection, code refinement, and multi-objective rewards~\citep{li2026brainpilot,wang2026neuroweaver}. 
NeuroAct extends the workflow with an EEG evidence loop that diagnoses gaps using signal, representation, predictor, and training evidence before selecting executable analyses or interventions.

\subsection{Experience Reuse}
Experience-transfer methods reuse outcomes across attempts or tasks. 
Reflexion and ExpeL reuse verbal feedback or contrasting trajectories, while ReasoningBank and EvoScientist distill and refine research strategies from outcomes~\citep{shinn2023reflexion,zhao2024expel,ouyang2026reasoningbank,lyu2026evoscientist}. 
Voyager and Recuris reuse executable skills or revise the memory governing their use, while SkillsBench and controlled studies examine when prescribed skills, tools, or guidance granularity hinder stronger strategies~\citep{wang2024voyager,yu2026recuris,li2026skillsbench,xu2026skill_granularity}. 
NeuroAct treats experience as a conditional, testable recommendation under explicit EEG task and model conditions, validating its use on new tasks through outcomes and cost before retaining, revising, or narrowing applicability.

\section{Method}
\label{sec:method}

\begin{figure}[t]
\centering
\includegraphics[width=\linewidth]{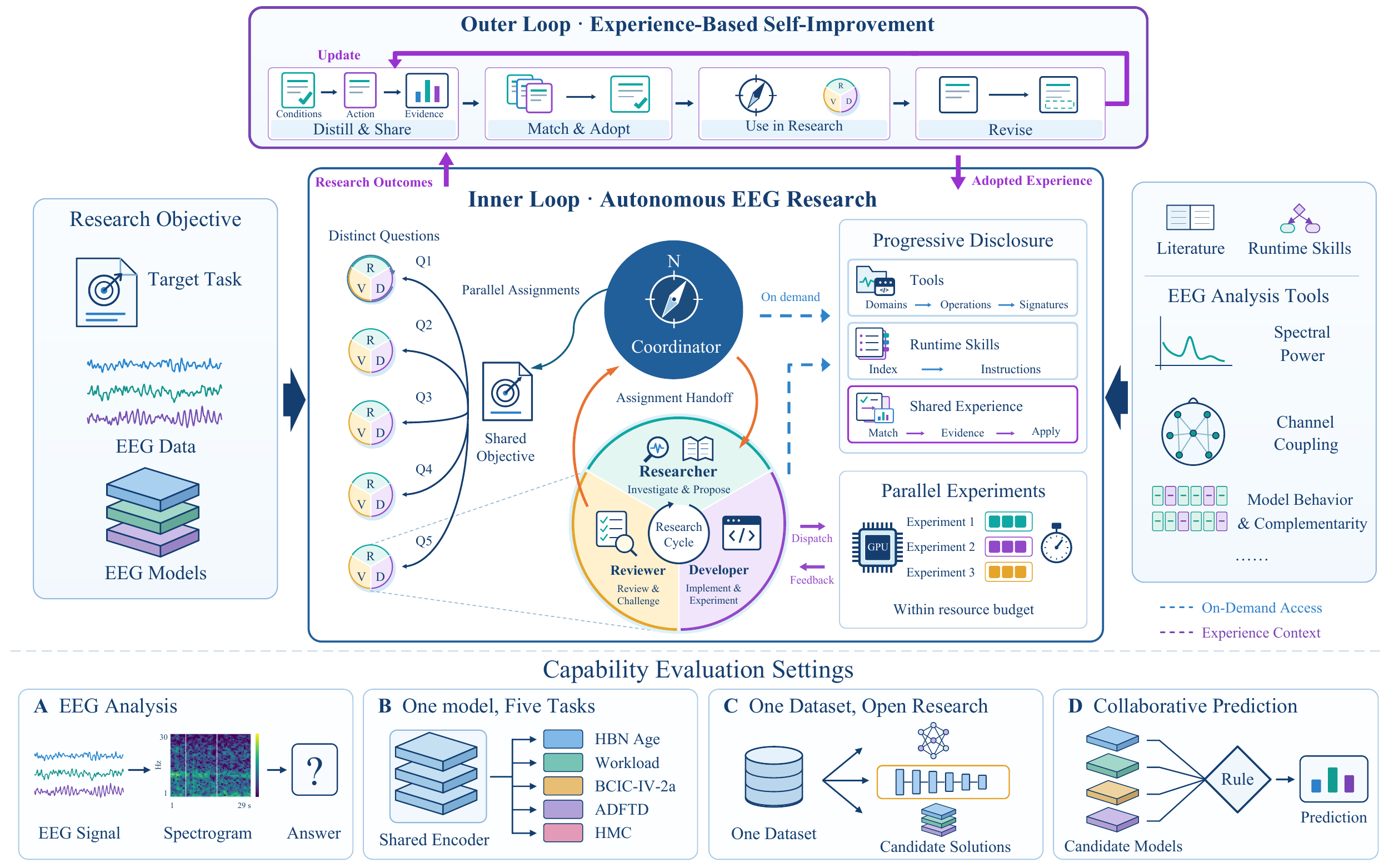}
\caption{NeuroAct's dual-loop autonomous research framework.
The Coordinator decomposes an EEG objective into parallel questions and directs separate cycles involving the Researcher, Developer, and Reviewer.
The inner loop selects actions from reviewed evidence and experimental feedback within budget.
The outer loop distills, matches, applies, and revises conditional experience across tasks.
Capabilities are disclosed on demand; the lower panels show four evaluation settings.}
\label{fig:neuroact_overview}
\end{figure}

NeuroAct organizes EEG analysis and decoding through two coupled research loops.
The inner loop~(Sec.~\ref{sec:Inner_Loop}) decomposes a shared objective into parallel questions, with the Coordinator directing specialists in separate contexts and selecting subsequent actions from reviewed evidence, experimental feedback, and remaining budget.
The outer loop~(Sec.~\ref{sec:Outer_Loop}) distills outcomes into conditional recommendations and revises them across tasks.
Fig.~\ref{fig:neuroact_overview} summarizes the framework; Alg.~\ref{alg:neuroact_research} specifies the interaction.
Sec.~\ref{sec:method_settings} presents four settings that exercise different capabilities.

\subsection{Inner Loop: Autonomous EEG Research}
\label{sec:Inner_Loop}
\noindent\textbf{Specialized collaboration.}
NeuroAct separates persona contexts and permissions so that scientific assessment, implementation, and review remain distinct.
The \textit{\textbf{Coordinator}} decomposes the objective into parallel questions and assigns each to a separate research cycle.
The \textbf{\textit{Researcher}} investigates an assigned question using literature, dataset semantics, and EEG evidence, then develops a feasible proposal.
The \textbf{\textit{Developer}} implements the selected proposal and executes the required analyses or training.
The \textbf{\textit{Reviewer}} examines the solution, implementation, and evidence, challenges unsupported conclusions, and returns the reviewed outcome to the Coordinator.
Each \emph{handoff} records the requested deliverable, supporting evidence, unresolved issues, and an actionable recommendation; the Coordinator integrates these handoffs, resolves disagreements, and selects subsequent actions.
\par\vspace{-\parskip}\noindent\textbf{Distinct investigations, coordinated decisions.}
For objective $g$, EEG data $\mathcal D$, and available model and code assets $\mathcal A$, the Coordinator selects a research decision $u_t$ and specialist assignments $\mathcal J_t$,
\begin{equation}
 (u_t,\mathcal J_t)=\pi_{\mathrm C}
 \bigl(g,\mathcal D,\mathcal A;\mathcal H_t,\mathcal P_t,\mathcal K_t,B_t\bigr),
 \label{eq:research_decision}
\end{equation}
where $\mathcal H_t$ records current-study evidence and decisions, $\mathcal P_t$ contains implemented solutions and results, $\mathcal K_t$ supplies applicable experience, and $B_t$ is the remaining budget.
The Coordinator can refine a candidate, redirect exploration, return to an earlier solution, or stop.
An assignment $(r_i,g_{t,i},\xi_{t,i})$ specifies a specialist role $r_i$ and question $g_{t,i}$, with task context $\xi_{t,i}$ containing relevant evidence, adopted experience, and the requested deliverable. The specialist returns
\begin{equation}
 h^r_{t,i}=\pi_r(g_{t,i},\xi_{t,i};\mathcal T_r),
 \qquad r\in\{\mathrm R,\mathrm D,\mathrm V\},
 \label{eq:specialist_handoff}
\end{equation}
where $h^r_{t,i}$ is the handoff returned to the Coordinator by role $r$ for assignment $i$ at iteration $t$, $\mathcal T_r$ denotes that role's tools, and $\mathrm V$ denotes the Reviewer.
The policies are prompted rather than trained; conditioning on accumulated evidence and adopted experience lets research behavior evolve without LLM parameter updates.
The Coordinator updates $\mathcal H_{t+1}$ with handoffs, $\mathcal P_{t+1}$ with completed solutions, and $B_{t+1}$ from actual consumption.
Research and review can proceed alongside training, while separate experiments run concurrently when resources permit.
Distinct questions define parallel branches; the Coordinator compares reviewed outcomes and retains the best-supported solutions.
\\\noindent\textbf{Progressive disclosure.}
Role contexts initially receive compact capability indexes and expand selected details on demand.
Tools expose domains before operations and input signatures; runtime skills expose indexes before instructions; matched experience exposes summaries before supporting evidence.
Only selected capabilities and adopted advice enter specialist assignments, concentrating context on the current question while preserving access to literature, operational details, and evidence.
App.~\ref{app:system} details the tools, skills, and disclosure levels.
\par\vspace{-\parskip}\noindent\textbf{Investigation becomes an experiment.}
The Researcher proposes an analysis whose possible outcomes distinguish competing explanations and imply different actions.
Literature and evidence on signal structure, representations, predictor behavior, and training dynamics identify the uncertainty, a feasible intervention, and its expected effect.
When computation is required, the Researcher specifies it for the Coordinator to assign to the Developer.
The Developer executes it or implements scoped changes to input processing, pretrained-model adaptation, predictors, or training.
Analytical tasks return computed results with evidence-supported interpretations.
\par\vspace{-\parskip}\noindent\textbf{Feedback redirects research.}
The Coordinator compares observed results and resource consumption with expected effects before selecting the next investigation.
Reviewer challenges to an implementation or explanation can trigger further investigation.
Auxiliary metrics and training behavior distinguish improved discrimination from decision-rule changes or unstable fitting.
This evidence updates the solution, research state, and next decision.

\begin{algorithm}[t]
    \caption{NeuroAct's inner and outer research loops}
    \label{alg:neuroact_research}
    \begin{algorithmic}[1]
        \REQUIRE Objective $g$, data $\mathcal D$, assets $\mathcal A$, budget $B_0$, shared experience $\mathcal E_j$
        \STATE Set $t\gets0$; Initialize $\mathcal H_0,\mathcal P_0$
        \STATE Retrieve and synthesize $\mathcal E_j$ for Coordinator adoption into $\mathcal K_0$ (Sec.~\ref{sec:Outer_Loop}).
        \WHILE{further research is warranted within budget}
            \STATE Coordinator selects $(u_t,\mathcal J_t)$ using Eq.~(\ref{eq:research_decision}) and records its rationale and expected effect.
            \STATE Dispatch $\mathcal J_t$ in parallel where resources permit.
            \FOR{each completed assignment $(r_i,g_{t,i},\xi_{t,i})\in\mathcal J_t$}
                \STATE \textbf{if} $r_i=\mathrm R$, collect Researcher handoff $h^{\mathrm R}_{t,i}$.
                \STATE \textbf{if} $r_i=\mathrm D$, collect Developer handoff $h^{\mathrm D}_{t,i}$.
                \STATE \textbf{if} $r_i=\mathrm V$, collect Reviewer handoff $h^{\mathrm V}_{t,i}$.
            \ENDFOR
            \STATE Integrate completed handoffs into $\mathcal H_{t+1}$; add completed solutions to $\mathcal P_{t+1}$.
            \STATE Reassess adopted recommendations against new outcomes; refresh $\mathcal K_{t+1}$ and remaining budget $B_{t+1}$; advance $t$.
        \ENDWHILE
        \STATE Distill eligible outcomes and recommendations in $\mathcal E_j$ using Eq.~(\ref{eq:experience_update}), yielding $\mathcal E_{j+1}$ for later studies.
        \ENSURE Coordinator-selected analytical answer or prediction solution with supporting evidence
    \end{algorithmic}
\end{algorithm}


\subsection{Outer Loop: Experience-Based Self-Improvement}
\label{sec:Outer_Loop}
\noindent\textbf{Distillation from contrasting outcomes.}
The outer loop distills experience about which experimental actions help under particular task and model conditions.
The Coordinator compares an intervention's expected and observed effects with considered alternatives, producing a conditional recommendation,
\begin{equation}
 k=(c,a,e^{+},e^{-},b),
 \label{eq:conditional_experience}
\end{equation}
where $c$ specifies applicable task and model conditions, $a$ the recommended action, $e^{+}$ and $e^{-}$ supporting and conflicting evidence, and $b$ the conditions for reassessment.
\par\vspace{-\parskip}\noindent\textbf{Matching and adoption.}
Retrieval filters ineligible or incompatible entries and ranks candidates by task and model compatibility, research-goal overlap, source evidence, EEG paradigm, and class structure.
Similar recommendations are grouped into bounded summaries with source links.
The Coordinator compares this synthesis with the target model and training conditions, choosing adoption, a bounded applicability check, or rejection.
Candidate advice is shown first to the Coordinator; specialists receive only adopted content relevant to their assignments.
Active recommendations enter role contexts within a character budget, with full evidence accessible on demand.
\par\vspace{-\parskip}\noindent\textbf{Revision through subsequent research.}
Target outcomes can narrow a recommendation, revise its action, or suspend its use.
Matched trials assess transfer through performance and research cost.
Study $j$ both uses and updates shared experience,
\begin{equation}
 \mathcal K_{j,t}=\operatorname{Context}_{\mathrm C}
 (\mathcal E_j;g_j,\mathcal D_j,\mathcal A_j,\mathcal H_{j,t}),\qquad
 \mathcal E_{j+1}=\operatorname{Revise}_{\mathrm C}
 (\mathcal E_j;\mathcal H_j^{\mathrm{dev}}),
\label{eq:experience_update}
\end{equation}
where $\operatorname{Context}_{\mathrm C}$ combines retrieval with the Coordinator's applicability decision, while $\operatorname{Revise}_{\mathrm C}$ distills or revises recommendations from eligible development outcomes.
Revised experience changes decisions in Eq.~(\ref{eq:research_decision}), generating evidence for further revision.
Unlike interaction memory, this recursion improves the agent's research strategy by revising which experiments it recommends and when, based on subsequent outcomes~\citep{yu2026recuris}.

\subsection{Four Settings for EEG Research}
\label{sec:method_settings}
\noindent\textbf{EEG General Analysis (A).}
Given a question and EEG recordings, NeuroAct selects computations and sources to produce an evidence-supported answer.
This setting exercises analytical reasoning and the investigative capabilities used in subsequent decoding research.
\par\vspace{-\parskip}\noindent\textbf{Cross-task Model Optimization (B).}
Each program adapts one pretrained encoder across heterogeneous datasets using task-specific adapters and output modules.
NeuroAct can revise task sampling and loss contributions, then inspect the effects of shared updates on each target.
Joint classification and regression create coupled objectives requiring balancing predictive utility across tasks.
\par\vspace{-\parskip}\noindent\textbf{Single-task Open Research (C).}
Given a dataset and prediction objective, NeuroAct can refine a strong candidate or explore another model family, including pretrained and task-specific predictors.
This setting exercises open development guided by EEG evidence.
\par\vspace{-\parskip}\noindent\textbf{Collaborative Prediction (D).}
NeuroAct examines aligned errors or regression residuals for complementarity and develops a combination or routing program,
$
 \hat y=F_{\omega}\!\left(\{f_m(x)\}_{m\in\mathcal M},z(x)\right),
$
where $\mathcal M$ denotes members and $z(x)$ contains inference-time EEG features or outputs.
Research can modify $F_{\omega}$ and, when permitted, train additional members to outperform the strongest predictor.

\section{Experiments}
\label{sec:experiments}
\setcounter{topnumber}{2}
\textbf{Experimental setup.}
We evaluate how effectively NeuroAct links EEG analysis to predictive research and improves later studies through experience.
Four settings test the breadth of these capabilities; mechanism ablations and experience studies measure effects on prediction and research cost.
BrainBench~(Setting A) evaluates instruction-driven EEG analysis across foundational, sleep, neurocognitive, and physiological domains.
Settings B, C, and D cover Workload, BCIC-IV-2a, ADFTD, HMC, and HBN; PhysioMI supports experience and supplementary studies.
DeepSeek-V4 Flash is default LLM; GPT-5.6 Luna and Qwen3.8 Flash Next assess provider consistency, with CodeAct, BrainPilot, and BrainAgent as baseline agents.
For Settings B--D, we adapt the baselines' task and execution interfaces.
Tab.~\ref{tab:agent_framework_comparison} describes originally reported capabilities; Tab.~\ref{tab:v6_frameworks}(B-D) evaluates the adapted implementations.
We include BENDR, BIOT, LaBraM, EEGPT, CBraMod, CSBrain, and REVE in model pools for Setting D, with ShallowFBCSPNet added for ADFTD.
Classification uses balanced accuracy and AUROC (BA, and AUC in tables); HBN regression uses Pearson $r$ and prediction errors.
Protected test labels remain unavailable during dual-loop research.
The research budget for each study is capped at 50 million LLM tokens or 5 GPU hours.
Research costs include all attempted experiments.
\textbf{Apps.~\ref{app:data} and \ref{app:baselines} detail tasks, execution, and protocols; App.~\ref{app:full_results} reports member pools and complete metrics.}

\subsection{Main Results Across the Four Research Settings}
We evaluate diverse dataset, EEG model, and LLM configurations. Tab.~\ref{tab:v6_configurations} follows NeuroAct's outcomes and reference solutions; Tab.~\ref{tab:v6_frameworks} compares external agent frameworks.
\begin{table}[t]
\captionsetup{skip=6pt}\centering\normalsize
\setlength{\tabcolsep}{1.6pt}\setlength{\parskip}{0pt}
\caption{NeuroAct research outcomes. Setting D reports the best single member, equal-weight reference, and NeuroAct-developed solution from its member pool.}\label{tab:v6_configurations}
\begingroup
\textbf{(A) EEG General Analysis}\\[2pt]
\resizebox{\linewidth}{!}{%
\begin{tabular*}{1.111\linewidth}{@{\extracolsep{\fill}}lrrrrr@{}}
\toprule
LLM & Found. $\uparrow$ & Sleep $\uparrow$ & Neurocog. $\uparrow$ & Physiol. $\uparrow$ & Macro $\uparrow$ \\
\midrule
DeepSeek & \textbf{87.26} & \textbf{75.36} & \textbf{68.73} & \textbf{92.16} & \textbf{80.88} \\
Luna & 84.96 & 70.27 & 67.54 & 79.55 & 75.58 \\
Qwen & 78.67 & 72.20 & 65.89 & 84.27 & 75.26 \\
\bottomrule
\end{tabular*}}\par\endgroup\vspace{1pt}
\begingroup
\textbf{(B) Cross-task Model Optimization}\\[2pt]
\resizebox{\linewidth}{!}{%
\begin{tabular*}{1.111\linewidth}{@{\extracolsep{\fill}}llrrrrrrrrrr@{}}
\toprule
 &  & \multicolumn{2}{c}{Workload} & \multicolumn{2}{c}{BCIC-IV-2a} & \multicolumn{2}{c}{ADFTD} & \multicolumn{2}{c}{HMC} & \multicolumn{2}{c}{HBN} \\
\cmidrule(lr){3-4}\cmidrule(lr){5-6}\cmidrule(lr){7-8}\cmidrule(lr){9-10}\cmidrule(lr){11-12}
Model & Stage & BA $\uparrow$ & AUC $\uparrow$ & BA $\uparrow$ & AUC $\uparrow$ & BA $\uparrow$ & AUC $\uparrow$ & BA $\uparrow$ & AUC $\uparrow$ & $r\uparrow$ & MAE $\downarrow$ \\
\midrule
LaBraM & Initial & 50.00 & 0.598 & 28.65 & 0.565 & 35.28 & 0.555 & 69.95 & 0.917 & 0.653 & 2.009 \\
 & NeuroAct & \textbf{65.29} & \textbf{0.770} & \textbf{33.94} & \textbf{0.625} & \textbf{56.37} & \textbf{0.726} & \textbf{70.74} & \textbf{0.928} & \textbf{0.656} & \textbf{1.998} \\
REVE & Initial & 49.55 & 0.815 & 25.43 & 0.581 & \textbf{53.09} & \textbf{0.753} & 52.80 & 0.857 & 0.616 & 2.149 \\
 & NeuroAct & \textbf{76.55} & \textbf{0.850} & \textbf{50.26} & \textbf{0.832} & 51.98 & 0.749 & \textbf{66.02} & \textbf{0.918} & \textbf{0.719} & \textbf{1.810} \\
EEGPT & Initial & 49.55 & 0.679 & 31.60 & 0.616 & \textbf{46.51} & 0.647 & 62.33 & 0.901 & 0.689 & 1.908 \\
 & NeuroAct & \textbf{68.00} & \textbf{0.735} & \textbf{34.03} & \textbf{0.710} & 45.98 & \textbf{0.659} & \textbf{71.35} & \textbf{0.932} & \textbf{0.736} & \textbf{1.749} \\
CBraMod & Initial & 50.00 & 0.624 & 28.82 & 0.624 & 46.05 & \textbf{0.732} & 60.29 & 0.887 & 0.624 & 2.058 \\
 & NeuroAct & \textbf{53.93} & \textbf{0.763} & \textbf{50.69} & \textbf{0.778} & \textbf{52.54} & 0.720 & \textbf{69.95} & \textbf{0.918} & \textbf{0.633} & \textbf{1.962}  \\
\bottomrule
\end{tabular*}}\par\endgroup\vspace{1pt}
\begingroup
\textbf{(C) Single-task Open Research}\\[2pt]
\resizebox{\linewidth}{!}{%
\begin{tabular*}{1.111\linewidth}{@{\extracolsep{\fill}}lrrrrrrrrrr@{}}
\toprule
 & \multicolumn{2}{c}{Workload} & \multicolumn{2}{c}{BCIC-IV-2a} & \multicolumn{2}{c}{ADFTD} & \multicolumn{2}{c}{HMC} & \multicolumn{2}{c}{HBN} \\
\cmidrule(lr){2-3}\cmidrule(lr){4-5}\cmidrule(lr){6-7}\cmidrule(lr){8-9}\cmidrule(lr){10-11}
Stage & BA $\uparrow$ & AUC $\uparrow$ & BA $\uparrow$ & AUC $\uparrow$ & BA $\uparrow$ & AUC $\uparrow$ & BA $\uparrow$ & AUC $\uparrow$ & $r\uparrow$ & MAE $\downarrow$ \\
\midrule
Compared & 81.92 & 0.884 & 52.60 & 0.789 & 54.93 & \textbf{0.751} & 73.03 & 0.934 & 0.488 & 2.425 \\
NeuroAct & \textbf{83.42} & \textbf{0.904} & \textbf{60.59} & \textbf{0.839} & \textbf{71.26} & 0.748 & \textbf{73.36} & \textbf{0.938} & \textbf{0.715} & \textbf{1.791} \\
\bottomrule
\end{tabular*}}\par\endgroup\vspace{1pt}
\begingroup
\textbf{(D) Collaborative Prediction}\\[2pt]
\resizebox{\linewidth}{!}{%
\begin{tabular*}{1.111\linewidth}{@{\extracolsep{\fill}}lrrrrrrrrrr@{}}
\toprule
 & \multicolumn{2}{c}{Workload} & \multicolumn{2}{c}{BCIC-IV-2a} & \multicolumn{2}{c}{ADFTD} & \multicolumn{2}{c}{HMC} & \multicolumn{2}{c}{HBN} \\
\cmidrule(lr){2-3}\cmidrule(lr){4-5}\cmidrule(lr){6-7}\cmidrule(lr){8-9}\cmidrule(lr){10-11}
Predictor & BA $\uparrow$ & AUC $\uparrow$ & BA $\uparrow$ & AUC $\uparrow$ & BA $\uparrow$ & AUC $\uparrow$ & BA $\uparrow$ & AUC $\uparrow$ & $r\uparrow$ & MAE $\downarrow$ \\
\midrule
Best single & 79.69 & 0.888 & 60.85 & \textbf{0.844} & 71.26 & 0.748 & 73.12 & 0.934 & 0.758 & 1.661 \\
Equal weights & 77.95 & 0.895 & 58.07 & 0.837 & 64.44 & 0.799 & 75.02 & 0.943 & 0.738 & 1.757 \\
NeuroAct & \textbf{85.02} & \textbf{0.897} & \textbf{61.02} & 0.841 & \textbf{74.89} & \textbf{0.843} & \textbf{75.14} & \textbf{0.945} & \textbf{0.780} & \textbf{1.585} \\
\bottomrule
\end{tabular*}}\par\endgroup\vspace{1pt}
\end{table}
\\\textbf{EEG General Analysis across domains.}
In Tab.~\ref{tab:v6_configurations}(A), NeuroAct achieves an 80.88 macro score with DeepSeek and exceeds 75 with both Luna and Qwen, showing consistent analytical capability across providers. DeepSeek attains 87.26 in foundational analysis and 92.16 in physiological integration, demonstrating effective use of EEG computations across distinct analytical objectives. Tab.~\ref{tab:v6_frameworks}(A) compares these outcomes with external agents, with NeuroAct leading every reported domain.
\\\textbf{Cross-task Model Optimization balances heterogeneous target tasks.}
Tab.~\ref{tab:v6_configurations}(B) examines joint classification and regression with a shared encoder.
With EEGPT, NeuroAct raises four-target mean BA from 47.50\% to 54.84\% and HBN $r$ from 0.6894 to 0.7362, while reducing MSE from 6.080 to 5.447.
These gains within one shared checkpoint show its ability to balance heterogeneous targets.
The REVE and CBraMod studies extend shared-model research to different encoder families.
In the LaBraM comparison in Tab.~\ref{tab:v6_frameworks}(B), NeuroAct leads the external agents on every target, including 56.37\% ADFTD BA against the strongest external result of 33.33\%.
App.~\ref{app:full_results} gives full results.
\\\textbf{Single-task Open Research develops task-specific solutions.}
Given a dataset and objective, Setting C lets the agent consult literature, implement methods, and redirect research. Compared in Tab.~\ref{tab:v6_configurations}(C) restricts research to existing program interfaces without authoring new scientific code. On ADFTD, NeuroAct implements class-balanced training and explores a different architecture, improving BA from 54.93\% with EEGNet to 71.26\% with ShallowFBCSPNet. On BCIC-IV-2a, the REVE solution reaches 60.59\% BA and 0.839 AUROC, improving on the restricted reference's 52.60\% BA. Tab.~\ref{tab:v6_frameworks}(C) shows a 15.36-point BCIC-IV-2a gain over the strongest completed external result. These results show the value of revising the predictive approach as evidence accumulates. Metrics appear in App.~\ref{app:full_results}.
\\\textbf{Collaborative Prediction turns complementarity into usable gain.}
Tab.~\ref{tab:v6_configurations}(D) compares NeuroAct with individual predictors and averaging.
On Workload, averaging ranks samples well; shifting pooled scores using training class proportions raises BA to 85.02\%, exceeding the best member's 79.69\%.
On HBN, convex weights shrunk toward equal weighting improve Pearson $r$ from 0.758 to 0.780 and reduce MAE from 1.661 to 1.585 over the best member.
NeuroAct leads external agents on Workload in Tab.~\ref{tab:v6_frameworks}(D).
On ADFTD, sample-dependent combination improves BA from the best member's 71.26\% to 74.89\% and AUROC from 0.748 to 0.843. App.~\ref{app:interventions} details these methods.

\begin{table}[t]
\captionsetup{skip=6pt}\centering
\caption{Agent comparison with DeepSeek. Setting A includes source-reported CodeAct and BrainAgent scores~\citep{zhou2026brainbench}. Setting B uses LaBraM. Settings B, C, and D report BA (\%) and HBN Pearson $r$. Metrics are detailed in App.~\ref{app:full_results}.}\label{tab:v6_frameworks}
\footnotesize\setlength{\tabcolsep}{1.15pt}
\begin{minipage}[t]{0.49\linewidth}\vspace{0pt}
\textbf{(A) EEG General Analysis}\par
\begin{tabular*}{\linewidth}{@{\extracolsep{\fill}}lrrrrr@{}}
\toprule
Agent & Found. & Sleep & Neuro. & Phys. & Macro \\
\midrule
CodeAct & 68.16 & 61.01 & 60.38 & 54.52 & 61.02 \\
BrainAgent & 67.38 & 67.97 & 67.21 & 49.98 & 63.13 \\
BrainPilot & 64.82 & 58.43 & 50.83 & 61.52 & 58.90 \\
NeuroAct & \textbf{87.26} & \textbf{75.36} & \textbf{68.73} & \textbf{92.16} & \textbf{80.88} \\
\bottomrule
\end{tabular*}\end{minipage}
\hfill\begin{minipage}[t]{0.49\linewidth}\vspace{0pt}
\textbf{(B) Cross-task Model Optimization: LaBraM}\par
\begin{tabular*}{\linewidth}{@{\extracolsep{\fill}}lrrrrr@{}}
\toprule
Agent & Work. & BCIC-IV-2a & ADFTD & HMC & HBN \\
\midrule
CodeAct & 50.00 & 25.00 & 32.39 & 39.30 & 0.228 \\
BrainAgent & 50.00 & 27.78 & 33.33 & 36.34 & 0.334 \\
BrainPilot & 49.55 & 28.12 & 24.82 & 60.30 & 0.500 \\
NeuroAct & \textbf{65.29} & \textbf{33.94} & \textbf{56.37} & \textbf{70.74} & \textbf{0.656} \\
\bottomrule
\end{tabular*}\end{minipage}
\par\smallskip
\begin{minipage}[t]{0.49\linewidth}\vspace{0pt}
\textbf{(C) Single-task Open Research}\par
\begin{tabular*}{\linewidth}{@{\extracolsep{\fill}}lrrrrr@{}}
\toprule
Agent & Work. & BCIC-IV-2a & ADFTD & HMC & HBN \\
\midrule
CodeAct & 60.11 & 45.23 & 56.60 & 72.90 & 0.687 \\
BrainAgent & 83.04 & 41.75 & 70.51 & 48.09 & 0.559 \\
BrainPilot & 64.54 & 40.45 & 62.25 & 71.46 & 0.547 \\
NeuroAct & \textbf{83.42} & \textbf{60.59} & \textbf{71.26} & \textbf{73.36} & \textbf{0.715} \\
\bottomrule
\end{tabular*}\end{minipage}
\hfill\begin{minipage}[t]{0.49\linewidth}\vspace{0pt}
\textbf{(D) Collaborative Prediction}\par
\begin{tabular*}{\linewidth}{@{\extracolsep{\fill}}lrrrrr@{}}
\toprule
Agent & Work. & BCIC-IV-2a & ADFTD & HMC & HBN \\
\midrule
CodeAct & 77.58 & 35.50 & 58.28 & 74.46 & 0.760 \\
BrainAgent & 76.44 & 60.68 & 52.61 & 74.50 & 0.758 \\
BrainPilot & 76.44 & 60.59 & 62.04 & 73.91 & 0.767 \\
NeuroAct & \textbf{85.02} & \textbf{61.02} & \textbf{74.89} & \textbf{75.14} & \textbf{0.780} \\
\bottomrule
\end{tabular*}\end{minipage}
\par\smallskip
\end{table}

\subsection{Ablations and Experience-Based Improvement}
\label{sec:experiment_mechanisms}

\begin{table}[t]
\captionsetup{skip=3pt}\centering\footnotesize
\setlength{\tabcolsep}{2pt}\setlength{\parskip}{0pt}
\caption{Mechanism ablations with DeepSeek and a fixed CBraMod family. Full reports the best complete-framework BA within this model constraint. App.~\ref{app:v6_selection} gives complete metrics.}\label{tab:v6_ablations}
\begin{tabular*}{\linewidth}{@{\extracolsep{\fill}}lrrrrrr@{}}
\toprule
 & \multicolumn{2}{c}{Workload} & \multicolumn{2}{c}{BCIC-IV-2a} & \multicolumn{2}{c}{ADFTD} \\
\cmidrule(lr){2-3}\cmidrule(lr){4-5}\cmidrule(lr){6-7}
Ablation & BA $\uparrow$ & AUC $\uparrow$ & BA $\uparrow$ & AUC $\uparrow$ & BA $\uparrow$ & AUC $\uparrow$ \\
\midrule
Full NeuroAct & \textbf{77.73} & \textbf{0.886} & \textbf{56.34} & \textbf{0.805} & \textbf{62.94} & \textbf{0.831} \\
w/o specialist personas & 77.11 & 0.853 & 51.65 & 0.778 & 61.29 & 0.723 \\
w/o EEG diagnostics & 72.63 & 0.752 & 52.17 & 0.798 & 41.80 & 0.579 \\
w/o code editing & 66.22 & 0.722 & 52.60 & 0.789 & 48.70 & 0.576 \\
w/o runtime skills & 70.60 & 0.783 & 52.60 & 0.788 & 48.95 & 0.633 \\
w/o auxiliary metrics & 77.29 & 0.862 & 49.74 & 0.786 & 61.62 & 0.829 \\
\bottomrule
\end{tabular*}
\end{table}

\begin{figure}[t]
    \centering
    \includegraphics[width=0.86\linewidth]{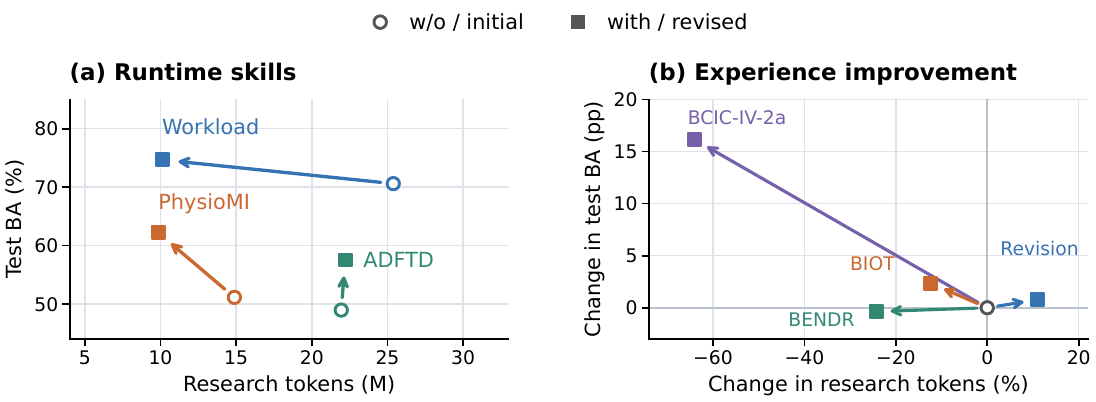}
    \caption{Performance-cost tradeoffs. (a) Absolute outcomes by runtime-skill condition. (b) Changes from each study's w/o or initial-advice reference at the origin. Token changes are percentages. Upward-leftward movement indicates better prediction at lower token cost.}
    \label{fig:v6_efficiency}
\end{figure}

\noindent\textbf{Ablating the research mechanisms.}
Tab.~\ref{tab:v6_ablations} fixes CBraMod and DeepSeek on three datasets to test the framework.
The ablations replace specialist collaboration with one persona, restrict EEG analysis to basic statistics, remove runtime skills, or limit code editing to configurable registered recipes rather than workspace implementations.
The auxiliary-metric ablation receives only the primary score; Full additionally receives macro-F1, AUROC, training loss, and gradient information.
Each ablation removes one capability while preserving the surrounding setup; App.~\ref{app:v6_selection} provides full metrics and costs.
Full NeuroAct improves Workload BA by 11.51 points over registered recipes, demonstrating the value of interventions beyond parameter selection.
Full exceeds the basic-statistics condition by 21.14 BA points on ADFTD and the primary-metric-only condition by 6.60 points on BCIC-IV-2a.
Runtime skills improve BA across all three datasets.
Fig.~\ref{fig:v6_efficiency}(a) reports paired runtime-skill studies on Workload, ADFTD, and PhysioMI in App.~\ref{app:v6_selection}, showing better prediction with lower token use on Workload and PhysioMI and an ADFTD gain of 8.57 percentage points at similar token cost.
\\\textbf{Experience transfer and revision.}
Tab.~\ref{tab:v6_experience} defaults to CBraMod and tests experience in later research.
Cross-dataset transfer carries patch-information advice from PhysioMI to BCIC-IV-2a.
Cross-model transfer fixes PhysioMI and provides CBraMod-derived experience to BENDR or BIOT.
Each begins with or without the source recommendation.
The revision study uses new source evidence to update advice on interpreting training dynamics and planning experiments, then tests it in new research.
Updated advice raises PhysioMI BA from 62.36\% to 63.13\% and reduces research time from 149.8 to 139.1 minutes.
Cross-dataset reuse raises BCIC-IV-2a BA from 34.66\% to 50.81\% and cuts tokens from 29.21M to 10.51M.
BENDR has similar BA with 24.28\% fewer research tokens and higher AUROC.
BIOT gains 2.34 BA points with 12.34\% fewer tokens and 22.08\% less research time.
Fig.~\ref{fig:v6_efficiency}(b) distinguishes performance and cost effects.
App.~\ref{app:transfer} provides complete metrics.

\subsection{Research Decision Traces}
\label{sec:analysis}
\noindent\textbf{EEG analysis shapes an executable predictive method.}
In an ADFTD study, the Researcher examines the recording montage and investigates spectral features suited to the task. The absence of a declared reference makes absolute power difficult to compare, motivating within-window relative power and frequency-band ratios. The Developer implements these features with regional aggregation and a gradient-boosting classifier. Error analysis finds low recall for frontotemporal dementia (FTD), with nonzero recall for every class. The agent treats class separation as the next research question and tests complementary connectivity and spatial-covariance features. The spectral solution remains the strongest evaluated approach in this investigation, reaching 67.27\% BA and 0.811 AUROC. EEG-specific analysis thus informs both implementation and the next experiment.
\par\vspace{-\parskip}\noindent\textbf{Model feedback changes the intervention.}
Workload research illustrates two ways to act on model behavior. In one study, early overfitting and variable validation performance motivate training-time perturbations scaled to the EEG amplitude. The resulting CBraMod comparison increases BA from 65.98\% to 71.62\% and AUROC from 0.768 to 0.805. In another, lowering the encoder learning rate leaves validation BA unchanged at 70.44\%. Investigation of class imbalance then leads to a class-weighted objective, reaching 76.89\% validation BA and 75.55\% test BA. The traces illustrate how feedback can redirect an intervention toward a better-supported explanation.
\par\vspace{-\parskip}\noindent\textbf{Experience focuses the next experiment.}
The PhysioMI recommendation directs BCIC-IV-2a research toward retaining CBraMod patch information. The first implementation also changes the training recipe, so the Coordinator requests a narrower comparison that preserves the established recipe. The resulting solution reaches 50.81\% BA with 64.02\% fewer research tokens than the study without experience. Reuse thus provides a concrete research direction, while target-side inspection refines how it is tested. App.~\ref{app:cases} details the evidence and implementations behind these decisions.

\begin{table}[t]
\captionsetup{skip=3pt}
\centering
\footnotesize
\setlength{\tabcolsep}{2pt}\setlength{\parskip}{0pt}
\caption{Experience transfer and revision with DeepSeek. All rows report full research costs.}\label{tab:v6_experience}
\begin{tabular*}{\linewidth}{@{\extracolsep{\fill}}llrrrr@{}}
\toprule
Study & Experience & BA $\uparrow$ & AUC $\uparrow$ & Tokens (M) $\downarrow$ & Min $\downarrow$ \\
\midrule
PhysioMI $\to$ BCIC-IV-2a   & w/o   & 34.66 & 0.592 & 29.21 & 236.7 \\
                            & with  & \textbf{50.81} & \textbf{0.775} & \textbf{10.51} & \textbf{165.3} \\
CBraMod $\to$ BENDR         & w/o   & \textbf{58.65} & 0.806 & 29.00 & 264.1 \\
                            & with  & 58.28 & \textbf{0.811} & \textbf{21.96} & \textbf{219.2} \\
CBraMod $\to$ BIOT          & w/o   & 47.92 & 0.730 & 10.78 & 209.7 \\
                            & with  & \textbf{50.26} & \textbf{0.749} & \textbf{9.45} & \textbf{163.4} \\
PhysioMI advice revision    & initial & 62.36 & 0.840 & \textbf{13.90} & 149.8 \\
                            & revised & \textbf{63.13} & \textbf{0.847} & 15.43 & \textbf{139.1} \\
\bottomrule
\end{tabular*}
\end{table}

\section{Conclusion}
\label{sec:conclusion}
NeuroAct advances autonomous EEG analysis and decoding by making domain evidence part of experimental decision-making rather than treating adaptation as undirected search.
Its inner loop turns investigation into executable interventions, while the outer loop carries useful experience into later research and revises it with new evidence.
The four settings demonstrate EEG General Analysis, Cross-task Model Optimization, Single-task Open Research, and Collaborative Prediction.
Fixed-model ablations support the contribution of specialized research capabilities, and experience studies across datasets and model families show both predictive gains and reduced research cost.
These results establish a path toward EEG research agents that improve current solutions, accumulate validated recommendations across tasks, and conduct subsequent studies more effectively.

\section*{AI use statement}
AI was used primarily to run the research agents evaluated in this work and to assist with manuscript writing and language polishing. The authors reviewed and revised the manuscript and take full responsibility for its content.

\section*{Ethics statement}
This study reuses existing EEG datasets and does not recruit new participants. EEG recordings and associated demographic or clinical labels require appropriate privacy protections, and dataset use remains subject to the original access and licensing terms. We report aggregate benchmark results. NeuroAct is evaluated as a research framework; clinical use would require independent validation and assessment of demographic and recording-condition shifts.

\section*{Reproducibility statement}
Sec.~\ref{sec:method} describes NeuroAct's research procedure. App.~\ref{app:data} documents dataset views, prediction targets, and evaluation units; Apps.~\ref{app:system} and \ref{app:prompts} specify the runtime, tools, skills, and role instructions. Complete predictive metrics appear in App.~\ref{app:full_results}, while App.~\ref{app:baselines} details baseline adaptations, ablations, and experience studies. Apps.~\ref{app:interventions} and \ref{app:cases} describe implemented interventions and the evidence behind representative research decisions.

\begingroup\raggedright
\bibliography{iclr2027_conference}
\endgroup
\bibliographystyle{iclr2027_conference}

\clearpage
\appendix
\FloatBarrier
\section{Data and Prediction Tasks}
\label{app:data}

\subsection{Prediction targets and evaluation units}
HBN provides pediatric age regression from resting EEG~\citep{alexander2017hbn,shirazi2024hbn_eeg}; Workload distinguishes mental-arithmetic conditions~\citep{zyma2019eegmat}; PhysioMI~\citep{schalk2004bci2000} and BCIC-IV-2a~\citep{tangermann2012bci_competition} provide four-class motor imagery; ADFTD distinguishes Alzheimer's disease, frontotemporal dementia, and healthy controls~\citep{miltiadous2023adftd}; HMC provides five-stage sleep classification~\citep{alvarez2021hmc}. Recordings are partitioned at the required subject or recording level before constructing prediction units.

\begin{table}[!htb]

\centering\small
\caption{Dataset characteristics and prediction-unit counts.}
\label{tab:v5_data}
\begin{tabular*}{\linewidth}{@{\extracolsep{\fill}}llrrrrr@{}}
\toprule
Dataset & Target & Channels & Window (s) & Training & Validation & Test\\
\midrule
HBN & Age regression & 129 & 2 & 121,740 & 28,620 & 19,740\\
Workload & Binary classification & 19 & 4 & 1,537 & 300 & 297\\
PhysioMI & 4-class classification & 64 & 4 & 6,210 & 1,734 & 1,893\\
ADFTD & 3-class classification & 19 & 10 & 4,743 & 1,115 & 1,155\\
HMC & 5-class classification & 4 & 30 & 91,681 & 22,804 & 22,440\\
BCIC-IV-2a & 4-class classification & 22 & 4 & 2,784 & 1,152 & 1,152\\
\bottomrule
\end{tabular*}
\end{table}

Training, validation, and test splits follow EEG-FM-Bench~\citep{xiong2026eegfmbench} for the classification datasets and NeuralBench~\citep{banville2026neuralbench} for HBN. Tab.~\ref{tab:v5_data} lists their sizes. HBN contributes 60 windows per subject, using R10 for validation and R5 for testing. 

HBN metrics are computed over windows, with MAE and RMSE in years and MSE in squared years. For classsification tasks, BA is mean class recall. Macro-F1 averages class F1 scores. AUROC uses the recorded binary or multiclass one-versus-rest calculation. Evaluation units and metric implementations are held fixed within each comparison.

\subsection{Model inputs and pretraining provenance}
Predictor inputs use 200 Hz, except EEGPT and LUNA at 256 Hz. EEGPT converts microvolt ($\mu$V) signals to millivolts (mV) by multiplying by 0.001; CBraMod, LaBraM, and CSBrain multiply $\mu$V values by 0.01. The ADFTD ShallowFBCSPNet uses $\mu$V signals directly.

BENDR maps each window to $[-1,1]$ using its minimum and maximum over all channels and time points. BIOT divides each channel by the 95th percentile of its absolute values within the window. EEGNet and LUNA apply channel-wise window z-scores. REVE also uses channel-wise window z-scores and clips values to $[-15,15]$. Its HBN data undergo channel-wise standardization over complete recordings before windowing~\citep{banville2026neuralbench}; candidate settings either preserve these values or apply an additional window-level z-score.

LaBraM and EEGPT include PhysioMI in their pretraining corpora~\citep{jiang2024labram,wang2024eegpt}. Main comparisons for Settings B, C, and D and the main CBraMod ablations therefore use BCIC-IV-2a for motor imagery. REVE's pretraining corpus also includes HBN releases, creating dataset overlap for its HBN evaluations and collaborative predictors containing REVE~\citep{elouahidi2025reve}. Following NeuralBench~\citep{banville2026neuralbench}, we retain HBN to evaluate agent-driven improvements under pretrained initializations with known data overlap.

\subsection{BrainBench comparison conditions}
The official BrainBench release contains 950 Foundational Analysis, 1,025 Sleep Assessment, 1,035 Neurocognitive Assessment, and 1,095 Physiological Integration cases~\citep{zhou2026brainbench}. Difficulty weighting is applied within subsets, followed by an equal average of their scores. Execution errors remain in the denominator.

The CodeAct and BrainAgent scores in Tab.~\ref{tab:v6_frameworks}(A) are source-reported BrainBench results~\citep{zhou2026brainbench}; BrainPilot and NeuroAct scores come from our evaluations. App.~\ref{app:baselines} describes the execution configurations.

\FloatBarrier
\section{NeuroAct Harness and Implementation}
\label{app:system}

\subsection{Runtime, roles, and execution}
NeuroAct implements the dual-loop framework using the OpenHands Software Agent SDK~\citep{wang2025openhandssoftwareagentsdk} and tools version 1.42.1. Its Agent and Conversation abstractions handle LLM interaction and tool execution. NeuroAct builds the research harness on this foundation, coordinating specialists and providing EEG analysis, literature tools, and executable model experiments. It manages evaluation and research evidence, with progressive disclosure of tools and runtime skills and support for experience reuse and revision. All four personas use one runtime definition, configured with shared base instructions and role-specific prompts and permissions. Shell access and workspace editing are assigned to the Developer. Researcher and Reviewer use read-only tools; the Reviewer can inspect the Developer's code changes together with run and evaluation records. Tool access follows the role bound to each conversation, keeping these permissions consistent as assignments change.

To make concurrent investigations complementary, the Coordinator specifies distinct questions and investigative paths before marking independent Researcher and Reviewer assignments for parallel execution. Each receives its own conversation and resource allowance, including concurrent assignments to the same role. Sequential assignments can reuse a compatible conversation to preserve useful context. The coordination host runs authorized assignments within the concurrency limit and forwards specialist handoffs to the Coordinator. This provides separate reasoning contexts for competing approaches while retaining a common point for their scientific assessment.

NeuroAct initializes a workspace at a user-specified location and uses bubblewrap to isolate the Developer's terminal, mounting research code as writable and reference resources as read-only. A submission specifies the implementation and its task and input configuration. Before training, preflight checks verify input shapes, channel and unit requirements, and executable interfaces. Submitted jobs run under the experiment service, which manages progress and cancellation or resumption independently of LLM turns. The experiment host returns execution feedback to the Developer's conversation. The Developer inspects the results and reports its findings in a handoff to the Coordinator. Run summaries expose training behavior and evaluation aggregates alongside resource use, supplying the concrete evidence needed for subsequent review and research decisions.

\subsection{Tools and progressive disclosure}
The runtime exposes four tool families, \texttt{neuro\_act\_research}, \texttt{neuro\_act\_experiment}, \texttt{neuro\_act\_knowledge}, and \texttt{neuro\_act\_collaboration}. The research family groups workspace, resource, literature, and analysis capabilities. Initial context contains a compact capability index. The agent obtains the selected callable contract before invocation, then requests detailed outputs when needed. Known capabilities can also be accessed directly. Run histories begin with compact summaries, with selected curves and artifacts retrieved on demand.

\begin{table}[!htb]
\centering\captionsetup{skip=5pt}
\caption{Research capabilities and their inputs and outputs. Access is filtered by persona.}\label{tab:v5_tools}
\small
\begin{tabular}{@{}>{\raggedright\arraybackslash}p{0.230\linewidth}>{\raggedright\arraybackslash}p{0.340\linewidth}>{\raggedright\arraybackslash}p{\dimexpr0.43\linewidth-4\tabcolsep\relax}@{}}
\toprule
Capability & Principal input & Returned evidence or operation\\
\midrule
Task and workspace context & Current task or selected candidate & Data/model specifications, research state, run summaries and selected detail\\
Literature and resources & Research query or named source & Retrieved papers or implementations, source content and compatibility information\\
EEG signal analysis & Recordings, event annotations and analysis configuration & Signal statistics and temporal, spectral or spatial structure, including event responses and connectivity\\
Model analysis & Candidate, run or prediction outputs & Input checks, optimization evidence, errors, perturbations and complementarity\\
Custom analysis & Developer-authored program and permitted inputs & Executed measurements with associated artifacts\\
Experiment execution & Implemented candidate and resource allowance & Training or prediction job, progress, outcomes and measured resource use\\
Knowledge and experience & Evidence-backed finding or target context & Scoped knowledge, ranked experience, adoption and revision records\\
Research collaboration & Bounded assignment or returned handoff & Specialist work, research decision and next authorized action\\
\bottomrule
\end{tabular}
\end{table}

NeuroAct's EEG analysis tools provide signal summaries and configurable measurements for specific research questions. Spectral analysis uses Welch estimates and configurable frequency bands. Spatial analysis summarizes covariance and channel relationships. Domain-shift analysis compares permitted partitions. Computed signal descriptors include band power, Hjorth measures, spectral entropy, peak frequency, and regional summaries; event-response and connectivity measurements are available for relevant tasks. Model analyses inspect input adaptation, gradients, optimization behavior, representations, perturbations, prediction errors, and complementary outputs. The agent selects analyses according to the current research question.

\subsection{Runtime skills}
Tab.~\ref{tab:v5_skills} lists the 16 skill triggers. Initial context exposes persona-filtered skill names and short trigger descriptions; the selected skill body is retrieved on demand. Each skill provides conditional guidance and stopping criteria to inform the agent's choice of experiments and implementations. Skills are shared research instructions; task-derived experience is updated from observed outcomes.

\begin{table}[!htb]
\centering\captionsetup{skip=5pt}
\caption{EEG runtime skills and the situations that trigger them.}\label{tab:v5_skills}
\small
\begin{tabular}{@{}p{0.300\linewidth}p{\dimexpr0.70\linewidth-2\tabcolsep\relax}@{}}
\toprule
Skill & Trigger\\
\midrule
Source investigation & An unfamiliar paper, model implementation or data source is encountered.\\
Dataset onboarding & The study introduces new data or unclear recording and label semantics.\\
Model onboarding & A model is unregistered, or its implementation needs revision.\\
EEG signal characterization & The study examines EEG properties or a signal-related bottleneck.\\
Sleep recording analysis & The question involves sleep stages or events in a sleep recording.\\
Event and behavioral analysis & The task links EEG responses to events or behavioral measurements.\\
Auxiliary signal analysis & Auxiliary physiological signals are analyzed alone or together with EEG.\\
Foundation-model adaptation & A pretrained EEG model is selected for the target task.\\
Training-pipeline design & The candidate requires a new training or fine-tuning procedure.\\
Iterative model improvement & A promising model-dataset pair still has a specific bottleneck.\\
Performance optimization & The task, evaluation protocol and budget are fixed for model optimization.\\
Specialist-model scouting & The task paradigm is known and further model candidates are needed.\\
Experiment diagnosis & Training, validation, prediction or resource behavior is unexpected.\\
Ensemble and routing & Multiple successful candidates show potentially complementary errors.\\
Candidate portfolio & The next step is to explore a new approach or refine a leading candidate.\\
Evidence and stopping & Candidate selection or a decision to continue or stop is needed.\\
\bottomrule
\end{tabular}
\end{table}

Most skills are available to Researcher, Developer, and Reviewer. Candidate portfolio is available to Coordinator and Researcher, and evidence and stopping to Coordinator, Researcher, and Reviewer. This access follows each persona's research responsibility. App.~\ref{app:prompts} reproduces role instructions and a representative skill excerpt.

\subsection{Experience formation, reuse, and revision}
\noindent\textbf{Formation from research records.}
To form an experience, the knowledge service assembles the decision rationale and expected effect with linked analyses and experiment outcomes. The Coordinator examines the implemented intervention and the alternatives that informed it, then specifies when to recommend the action and what would warrant reassessment. Each entry links supporting and conflicting evidence to its source records. The shared catalog indexes these entries by source task and model characteristics for retrieval across workspaces.

\noindent\textbf{Retrieval and synthesis.}
Retrieval first assesses compatibility between the source and target prediction tasks and model conditions, then ranks recommendations by a deterministic relevance score. Exact model agreement contributes 5 points and EEG-paradigm agreement 5; matching prediction type and dataset each contribute 3. Prediction and generalization units and input characteristics refine the score, while goal and analysis-term overlap adds up to 4 points. Results expose matched conditions and potential incompatibilities for the Coordinator's assessment. Repeated descriptions of the same source evidence are deduplicated, then entries are grouped by source task and model conditions and action terms.

\noindent\textbf{Adoption and task context.}
The Coordinator reviews the synthesis and relevant source details before assigning adopted recommendations to target tasks with a context priority. Provisional recommendations receive lower priority and a proposed low-cost applicability check. The runtime exposes adopted experience to the Coordinator and specialist contexts for the applicable tasks, grouped by action theme and ordered by priority. A default 6,000-character budget bounds this context summary, which retains actions and their conditions with references for on-demand access to complete entries. Long entries are abbreviated across their fields to retain the essential actions. When an experience changes a research decision, the Coordinator links the decision to the recommendation and revision used, connecting the subsequent outcome to that advice.

\noindent\textbf{Revision from subsequent evidence.}
Target results can lead the Coordinator to amend an adopted recommendation's action or applicability, update its priority, or suspend its use, with supporting observations and counterevidence attached to the revision. These task-specific updates retain the original source recommendation. To improve the shared recommendation itself, the Coordinator can reassess its original evidence or incorporate later outcomes, recording a revised recommendation with the reason for the change and links to preceding evidence. Context refreshes propagate active revisions to subsequent decisions and exclude withdrawn or superseded recommendations.

\subsection{Hardware and software environment}
Local experiments use an NVIDIA RTX 4070 Ti SUPER with 16 GB VRAM, an AMD Ryzen 5 9600X, and 96 GiB RAM.

The local software environment uses Ubuntu 24.04 LTS and Python 3.12. PyTorch 2.13.0 with CUDA 13.0 provides GPU computation; MNE-Python 1.12.1 and Braindecode 1.6.1 support EEG processing and model training. EEGDash 0.8.4 and MOABB 1.5.0 support dataset access, with scikit-learn 1.9.0 providing conventional learning and evaluation routines. Dependencies are managed with uv and pinned in the project environment.

\FloatBarrier
\section{Role Prompts and EEG Research Guidance}
\label{app:prompts}

The following excerpts reproduce the current English role instructions and their assigned research responsibilities.

\paragraph{Coordinator.}
\begin{quote}\small
You are NeuroAct's CoordinatorAgent: the multi-persona research lead and owner of continue, broaden, return, freeze, and stop decisions.

Separate candidate generation from winner selection.

Run a performance experiment directly when existing evidence identifies the action and signal. When unresolved mechanisms would lead to different actions, first run a discriminative analysis or experiment.
\end{quote}

\paragraph{Researcher.}
\begin{quote}\small
You are NeuroAct's ResearcherAgent. Investigate the frozen EEG task, dataset semantics, benchmark comparability, model assumptions, validation evidence, and competing mechanisms.

Choose research that changes a candidate, diagnosis, evaluation, fusion, or stopping decision.

Hand DeveloperAgent a minimal actionable plan with expected signal, alternative, source, and comparability limits.
\end{quote}

\paragraph{Developer.}
\begin{quote}\small
You are NeuroAct's DeveloperAgent. Own UserWorkspace code changes, custom analyses, candidate drafting, preflight, and experiment submission for the bounded delegation.

Implement an evidence-supported delta; do not reduce architecture, representation, loss, training, or routing evidence to cosmetic overrides.

Complete every delegation with observed evidence and interpretation; add recommendations, alternatives, and uncertainties only when evidence supports them.
\end{quote}

\paragraph{Reviewer.}
\begin{quote}\small
You are NeuroAct's ReviewerAgent. Prioritize scientific validity, attainable performance, and decision-changing defects over style or bookkeeping.

Name the decision under review and the strongest plausible alternative.

Your handoff should make four points easy to find: verdict, decisive evidence, strongest counterargument or alternative, and the next action that follows if the Coordinator accepts the audit.
\end{quote}

\paragraph{Runtime skill.}
The candidate-portfolio skill guides the choice between exploring alternatives and refining a promising candidate. Two excerpts are translated below.
\begin{quote}\small
The portfolio orders and retains alternatives; it is not an exhaustive execution queue. The Researcher can hand off candidates with different mechanisms and specify their information value. The Coordinator can form bounded concurrent batches through separate decisions and assignments, while each Developer assignment authorizes one candidate method group.

Exploring a new direction should state which current judgment it may overturn. Continuing a leading direction should state the expected bottleneck and observable success condition.
\end{quote}
The Coordinator and specialists apply this guidance to current findings and relevant experience when choosing the next experiment.

\FloatBarrier
\section{Detailed Results for the Four Research Settings}
\label{app:full_results}
The following tables provide complete predictive metrics for the four research settings. Classification results report BA (\%), macro-F1, and AUROC; HBN regression reports Pearson $r$, MAE, and MSE.

\subsection{EEG General Analysis (A)}
Tab.~\ref{tab:app_analysis} reports BrainBench's four domain scores for the framework comparison and NeuroAct's three LLM backends. NeuroAct leads each domain with DeepSeek and achieves macro-averages above 75 across all three backends.

\par\medskip\noindent\begin{minipage}{\linewidth}
\captionsetup{type=table,hypcap=false,skip=5pt}\centering
\caption{BrainBench domain scores.}\label{tab:app_analysis}
\footnotesize\setlength{\tabcolsep}{3pt}
\begin{tabular*}{\linewidth}{@{\extracolsep{\fill}}llrrrrr@{}}
\toprule
Agent & LLM & Found. & Sleep & Neurocog. & Physiol. & Macro \\
\midrule
CodeAct & DeepSeek & 68.16 & 61.01 & 60.38 & 54.52 & 61.02 \\
BrainAgent & DeepSeek & 67.38 & 67.97 & 67.21 & 49.98 & 63.13 \\
BrainPilot & DeepSeek & 64.82 & 58.43 & 50.83 & 61.52 & 58.90 \\
NeuroAct & DeepSeek & 87.26 & 75.36 & 68.73 & 92.16 & 80.88 \\
NeuroAct & Luna & 84.96 & 70.27 & 67.54 & 79.55 & 75.58 \\
NeuroAct & Qwen & 78.67 & 72.20 & 65.89 & 84.27 & 75.26 \\
\bottomrule
\end{tabular*}\end{minipage}\par\medskip

\subsection{Cross-task Model Optimization (B)}
Each shared checkpoint serves four classification tasks and HBN age regression, allowing improvements across heterogeneous targets to be examined together. Tabs.~\ref{tab:app_shared_first}--\ref{tab:app_shared_hbn} compare initial and revised checkpoints across four encoder families, including the macro-F1 and regression-error metrics that complement the main results. Tab.~\ref{tab:app_shared_agents} reports the corresponding framework comparison with DeepSeek and LaBraM under the five-task research objective.

\par\medskip\noindent\begin{minipage}{\linewidth}
\centering\captionsetup{type=table,hypcap=false,skip=5pt}
\caption{Classification results from Cross-task Model Optimization with LaBraM and REVE.}\label{tab:app_shared_first}
\footnotesize\setlength{\tabcolsep}{3pt}
\begin{tabular*}{\linewidth}{@{\extracolsep{\fill}}lllrrr@{}}
\toprule
Model & Stage & Dataset & BA $\uparrow$ & F1 $\uparrow$ & AUROC $\uparrow$ \\
\midrule
LaBraM & Initial & Workload & 50.00 & 0.428 & 0.598 \\
LaBraM & Initial & BCIC-IV-2a & 28.65 & 0.194 & 0.565 \\
LaBraM & Initial & ADFTD & 35.28 & 0.367 & 0.555 \\
LaBraM & Initial & HMC & 69.95 & 0.675 & 0.917 \\
LaBraM & NeuroAct & Workload & \textbf{65.29} & \textbf{0.672} & \textbf{0.770} \\
LaBraM & NeuroAct & BCIC-IV-2a & \textbf{33.94} & \textbf{0.341} & \textbf{0.625} \\
LaBraM & NeuroAct & ADFTD & \textbf{56.37} & \textbf{0.555} & \textbf{0.726} \\
LaBraM & NeuroAct & HMC & \textbf{70.74} & \textbf{0.710} & \textbf{0.928} \\
REVE & Initial & Workload & 49.55 & 0.426 & 0.815 \\
REVE & Initial & BCIC-IV-2a & 25.43 & 0.128 & 0.581 \\
REVE & Initial & ADFTD & \textbf{53.09} & 0.471 & \textbf{0.753} \\
REVE & Initial & HMC & 52.80 & 0.541 & 0.857 \\
REVE & NeuroAct & Workload & \textbf{76.55} & \textbf{0.746} & \textbf{0.850} \\
REVE & NeuroAct & BCIC-IV-2a & \textbf{50.26} & \textbf{0.453} & \textbf{0.832} \\
REVE & NeuroAct & ADFTD & 51.98 & \textbf{0.491} & 0.749 \\
REVE & NeuroAct & HMC & \textbf{66.02} & \textbf{0.653} & \textbf{0.918} \\
\bottomrule
\end{tabular*}
\end{minipage}\par\medskip

\par\medskip\noindent\begin{minipage}{\linewidth}
\centering\captionsetup{type=table,hypcap=false,skip=5pt}
\caption{Classification results from Cross-task Model Optimization with EEGPT and CBraMod.}\label{tab:app_shared_second}
\footnotesize\setlength{\tabcolsep}{3pt}
\begin{tabular*}{\linewidth}{@{\extracolsep{\fill}}lllrrr@{}}
\toprule
Model & Stage & Dataset & BA $\uparrow$ & F1 $\uparrow$ & AUROC $\uparrow$ \\
\midrule
EEGPT & Initial & Workload & 49.55 & 0.426 & 0.679 \\
EEGPT & Initial & BCIC-IV-2a & 31.60 & 0.213 & 0.616 \\
EEGPT & Initial & ADFTD & \textbf{46.51} & 0.408 & 0.647 \\
EEGPT & Initial & HMC & 62.33 & 0.606 & 0.901 \\
EEGPT & NeuroAct & Workload & \textbf{68.00} & \textbf{0.635} & \textbf{0.735} \\
EEGPT & NeuroAct & BCIC-IV-2a & \textbf{34.03} & \textbf{0.268} & \textbf{0.710} \\
EEGPT & NeuroAct & ADFTD & 45.98 & \textbf{0.458} & \textbf{0.659} \\
EEGPT & NeuroAct & HMC & \textbf{71.35} & \textbf{0.723} & \textbf{0.932} \\
CBraMod & Initial & Workload & 50.00 & 0.428 & 0.624 \\
CBraMod & Initial & BCIC-IV-2a & 28.82 & 0.219 & 0.624 \\
CBraMod & Initial & ADFTD & 46.05 & 0.417 & \textbf{0.732} \\
CBraMod & Initial & HMC & 60.29 & 0.620 & 0.887 \\
CBraMod & NeuroAct & Workload & \textbf{53.93} & \textbf{0.536} & \textbf{0.763} \\
CBraMod & NeuroAct & BCIC-IV-2a & \textbf{50.69} & \textbf{0.492} & \textbf{0.778} \\
CBraMod & NeuroAct & ADFTD & \textbf{52.54} & \textbf{0.528} & 0.720 \\
CBraMod & NeuroAct & HMC & \textbf{69.95} & \textbf{0.697} & \textbf{0.918} \\
\bottomrule
\end{tabular*}
\end{minipage}\par\medskip

\par\medskip\noindent\begin{minipage}{\linewidth}
\captionsetup{type=table,hypcap=false,skip=5pt}\centering
\caption{HBN age regression for the same Cross-task Model Optimization checkpoints.}\label{tab:app_shared_hbn}
\footnotesize\setlength{\tabcolsep}{3pt}
\begin{tabular*}{\linewidth}{@{\extracolsep{\fill}}llrrr@{}}
\toprule
Model & Stage & $r\uparrow$ & MAE $\downarrow$ & MSE $\downarrow$ \\
\midrule
LaBraM & Initial & 0.653 & 2.009 & \textbf{7.032} \\
LaBraM & NeuroAct & \textbf{0.656} & \textbf{1.998} & 7.036 \\
REVE & Initial & 0.616 & 2.149 & 7.724 \\
REVE & NeuroAct & \textbf{0.719} & \textbf{1.810} & \textbf{6.001} \\
EEGPT & Initial & 0.689 & 1.908 & 6.080 \\
EEGPT & NeuroAct & \textbf{0.736} & \textbf{1.749} & \textbf{5.447} \\
CBraMod & Initial & 0.624 & 2.058 & 7.359 \\
CBraMod & NeuroAct & \textbf{0.633} & \textbf{1.962} & \textbf{7.138} \\
\bottomrule
\end{tabular*}\end{minipage}\par\medskip

\par\medskip\noindent\begin{minipage}{\linewidth}
\centering\captionsetup{type=table,hypcap=false,skip=5pt}
\caption{Cross-task Model Optimization framework comparisons with DeepSeek and LaBraM. HBN columns report $r$, MAE, and MSE.}\label{tab:app_shared_agents}
\footnotesize\setlength{\tabcolsep}{3pt}
\begin{tabular*}{\linewidth}{@{\extracolsep{\fill}}llrrr@{}}
\toprule
Dataset & Agent & BA or $r$ & F1 or MAE & AUROC or MSE \\
\midrule
Workload & CodeAct & 50.00 & 0.428 & 0.573 \\
 & BrainAgent & 50.00 & 0.428 & 0.557 \\
 & BrainPilot & 49.55 & 0.426 & 0.739 \\
 & NeuroAct & \textbf{65.29} & \textbf{0.672} & \textbf{0.770} \\
\addlinespace[2pt]
BCIC-IV-2a & CodeAct & 25.00 & 0.100 & 0.520 \\
 & BrainAgent & 27.78 & 0.206 & 0.520 \\
 & BrainPilot & 28.12 & 0.182 & 0.582 \\
 & NeuroAct & \textbf{33.94} & \textbf{0.341} & \textbf{0.625} \\
\addlinespace[2pt]
ADFTD & CodeAct & 32.39 & 0.213 & 0.318 \\
 & BrainAgent & 33.33 & 0.190 & 0.494 \\
 & BrainPilot & 24.82 & 0.261 & 0.347 \\
 & NeuroAct & \textbf{56.37} & \textbf{0.555} & \textbf{0.726} \\
\addlinespace[2pt]
HMC & CodeAct & 39.30 & 0.384 & 0.748 \\
 & BrainAgent & 36.34 & 0.325 & 0.754 \\
 & BrainPilot & 60.30 & 0.598 & 0.887 \\
 & NeuroAct & \textbf{70.74} & \textbf{0.710} & \textbf{0.928} \\
\addlinespace[2pt]
HBN & CodeAct & 0.228 & 2.912 & 11.93 \\
 & BrainAgent & 0.334 & 2.636 & 12.27 \\
 & BrainPilot & 0.500 & 2.340 & 8.797 \\
 & NeuroAct & \textbf{0.656} & \textbf{1.998} & \textbf{7.036} \\
\bottomrule
\end{tabular*}
\end{minipage}\par\medskip

\subsection{Single-task Open Research (C)}
The research objective fixes the dataset and prediction protocol while permitting model selection and implementation changes. Compared restricts research to existing program interfaces without authoring new scientific code, providing the reference for the open-research solutions in Tab.~\ref{tab:v6_configurations}(C). The tables below identify the resulting predictors and report their complete classification and regression metrics, followed by the external-framework comparisons. The BrainAgent Workload result in Tab.~\ref{tab:app_open_agents} uses spectral features normalized across unlabeled windows within each test subject.

\par\medskip\noindent\begin{minipage}{\linewidth}
\captionsetup{type=table,hypcap=false,skip=5pt}\centering
\caption{Compared and selected solutions in Single-task Open Research, with complete classification metrics.}\label{tab:app_open_classification}
\footnotesize\setlength{\tabcolsep}{3pt}
\begin{tabular*}{\linewidth}{@{\extracolsep{\fill}}llrrr@{}}
\toprule
Dataset & Evaluated solution & BA $\uparrow$ & F1 $\uparrow$ & AUROC $\uparrow$ \\
\midrule
Workload & Compared: REVE & 81.92 & 0.807 & 0.884 \\
 & Selected: REVE & \textbf{83.42} & \textbf{0.819} & \textbf{0.904} \\
BCIC-IV-2a & Compared: CBraMod & 52.60 & 0.523 & 0.789 \\
 & Selected: REVE & \textbf{60.59} & \textbf{0.599} & \textbf{0.839} \\
ADFTD & Compared: EEGNet & 54.93 & 0.499 & \textbf{0.751} \\
 & Selected: ShallowFBCSPNet & \textbf{71.26} & \textbf{0.735} & 0.748 \\
HMC & Compared: EEGPT & 73.03 & \textbf{0.736} & 0.934 \\
 & Selected: BENDR & \textbf{73.36} & 0.731 & \textbf{0.938} \\
\bottomrule
\end{tabular*}\end{minipage}\par\medskip

\par\medskip\noindent\begin{minipage}{\linewidth}
\captionsetup{type=table,hypcap=false,skip=5pt}\centering
\caption{HBN predictors from Single-task Open Research, with complete regression metrics.}\label{tab:app_open_hbn}
\footnotesize\setlength{\tabcolsep}{3pt}
\begin{tabular*}{\linewidth}{@{\extracolsep{\fill}}lrrr@{}}
\toprule
Predictor & $r\uparrow$ & MAE $\downarrow$ & MSE $\downarrow$ \\
\midrule
Compared: BIOT & 0.488 & 2.425 & 9.785 \\
Additional: LUNA & 0.586 & 2.765 & 13.10 \\
NeuroAct: REVE & \textbf{0.715} & \textbf{1.791} & \textbf{5.845} \\
\bottomrule
\end{tabular*}\end{minipage}\par\medskip

\par\medskip\noindent\begin{minipage}{\linewidth}
\captionsetup{type=table,hypcap=false,skip=5pt}\centering
\caption{Single-task Open Research classification results with DeepSeek.}\label{tab:app_open_agents}
\footnotesize\setlength{\tabcolsep}{3pt}
\begin{tabular*}{\linewidth}{@{\extracolsep{\fill}}lrrrrrr@{}}
\toprule
 & \multicolumn{3}{c}{Workload} & \multicolumn{3}{c}{BCIC-IV-2a} \\
\cmidrule(lr){2-4}\cmidrule(lr){5-7}
Agent & BA $\uparrow$ & F1 $\uparrow$ & AUC $\uparrow$ & BA $\uparrow$ & F1 $\uparrow$ & AUC $\uparrow$ \\
CodeAct & 60.11 & 0.605 & 0.780 & 45.23 & 0.448 & 0.696 \\
BrainAgent & 83.04 & 0.787 & 0.890 & 41.75 & 0.372 & 0.764 \\
BrainPilot & 64.54 & 0.643 & 0.784 & 40.45 & 0.375 & 0.673 \\
NeuroAct & \textbf{83.42} & \textbf{0.819} & \textbf{0.904} & \textbf{60.59} & \textbf{0.599} & \textbf{0.839} \\
\midrule
 & \multicolumn{3}{c}{ADFTD} & \multicolumn{3}{c}{HMC} \\
\cmidrule(lr){2-4}\cmidrule(lr){5-7}
Agent & BA $\uparrow$ & F1 $\uparrow$ & AUC $\uparrow$ & BA $\uparrow$ & F1 $\uparrow$ & AUC $\uparrow$ \\
CodeAct & 56.60 & 0.521 & 0.793 & 72.90 & 0.706 & 0.931 \\
BrainAgent & 70.51 & 0.693 & 0.813 & 48.09 & 0.451 & 0.760 \\
BrainPilot & 62.25 & 0.622 & \textbf{0.821} & 71.46 & 0.713 & 0.932 \\
NeuroAct & \textbf{71.26} & \textbf{0.735} & 0.748 & \textbf{73.36} & \textbf{0.731} & \textbf{0.938} \\
\bottomrule
\end{tabular*}
\end{minipage}\par\medskip

\par\medskip\noindent\begin{minipage}{\linewidth}
\captionsetup{type=table,hypcap=false,skip=5pt}\centering
\caption{Single-task Open Research HBN results with DeepSeek.}\label{tab:app_open_agents_hbn}
\footnotesize\setlength{\tabcolsep}{3pt}
\begin{tabular*}{\linewidth}{@{\extracolsep{\fill}}lrrr@{}}
\toprule
Agent & $r\uparrow$ & MAE $\downarrow$ & MSE $\downarrow$ \\
\midrule
CodeAct & 0.687 & 1.942 & 6.092 \\
BrainAgent & 0.559 & 2.209 & 7.962 \\
BrainPilot & 0.547 & 2.215 & 8.138 \\
NeuroAct & \textbf{0.715} & \textbf{1.791} & \textbf{5.845} \\
\bottomrule
\end{tabular*}
\end{minipage}\par\medskip

\subsection{Collaborative Prediction (D)}
The comparisons below examine how NeuroAct's collaborative solutions improve on individual members and equal weighting, then compare the resulting performance across agent frameworks. Full metrics show that the Workload BA gain is accompanied by higher macro-F1 and AUROC, while the HBN solution improves both correlation and prediction error over the best single member.

Setting D reports each framework's highest evaluated predictive score for each task among its recorded Collaborative Prediction studies. Member composition varies across framework studies; the best-single and equal-weight references in Tab.~\ref{tab:v6_configurations}(D) correspond to the reported NeuroAct solution. BCIC-IV-2a uses sample-level routing and reaches 61.02\% BA. ADFTD uses a learned sample-dependent convex combination, developed through the research trajectory in App.~\ref{app:adftd_routing_case}.

\par\medskip\noindent\begin{minipage}{\linewidth}
\captionsetup{type=table,hypcap=false,skip=5pt}\centering
\caption{Classification references and best evaluated NeuroAct solutions with DeepSeek.}\label{tab:app_collab_classification}
\footnotesize\setlength{\tabcolsep}{3pt}
\begin{tabular*}{\linewidth}{@{\extracolsep{\fill}}llrrr@{}}
\toprule
Dataset & Predictor & BA $\uparrow$ & F1 $\uparrow$ & AUROC $\uparrow$ \\
\midrule
Workload & Best single & 79.69 & 0.788 & 0.888 \\
 & Equal weights & 77.95 & 0.791 & 0.895 \\
 & NeuroAct & \textbf{85.02} & \textbf{0.822} & \textbf{0.897} \\
BCIC-IV-2a & Best single & 60.85 & 0.600 & \textbf{0.844} \\
 & Equal weights & 58.07 & 0.568 & 0.837 \\
 & NeuroAct & \textbf{61.02} & \textbf{0.601} & 0.841 \\
ADFTD & Best single & 71.26 & \textbf{0.735} & 0.748 \\
 & Equal weights & 64.44 & 0.651 & 0.799 \\
 & NeuroAct & \textbf{74.89} & 0.729 & \textbf{0.843} \\
HMC & Best single & 73.12 & 0.727 & 0.934 \\
 & Equal weights & 75.02 & 0.749 & 0.943 \\
 & NeuroAct & \textbf{75.14} & \textbf{0.750} & \textbf{0.945} \\
\bottomrule
\end{tabular*}\end{minipage}\par\medskip

\par\medskip\noindent\begin{minipage}{\linewidth}
\captionsetup{type=table,hypcap=false,skip=5pt}\centering
\caption{HBN age regression references and NeuroAct's collaborative solution with DeepSeek.}\label{tab:app_collab_hbn}
\footnotesize\setlength{\tabcolsep}{3pt}
\begin{tabular*}{\linewidth}{@{\extracolsep{\fill}}lrrr@{}}
\toprule
Predictor & $r\uparrow$ & MAE $\downarrow$ & MSE $\downarrow$ \\
\midrule
Best single & 0.758 & 1.661 & 5.158 \\
Equal weights & 0.738 & 1.757 & 5.452 \\
NeuroAct & \textbf{0.780} & \textbf{1.585} & \textbf{4.702} \\
\bottomrule
\end{tabular*}\end{minipage}\par\medskip

\par\medskip\noindent\begin{minipage}{\linewidth}
\captionsetup{type=table,hypcap=false,skip=5pt}\centering
\caption{Collaborative Prediction classification results with DeepSeek.}\label{tab:app_collab_agents}
\footnotesize\setlength{\tabcolsep}{3pt}
\begin{tabular*}{\linewidth}{@{\extracolsep{\fill}}lrrrrrr@{}}
\toprule
 & \multicolumn{3}{c}{Workload} & \multicolumn{3}{c}{BCIC-IV-2a} \\
\cmidrule(lr){2-4}\cmidrule(lr){5-7}
Agent & BA $\uparrow$ & F1 $\uparrow$ & AUC $\uparrow$ & BA $\uparrow$ & F1 $\uparrow$ & AUC $\uparrow$ \\
CodeAct & 77.58 & 0.710 & 0.819 & 35.50 & 0.309 & 0.667 \\
BrainAgent & 76.44 & 0.690 & 0.802 & 60.68 & 0.596 & 0.829 \\
BrainPilot & 76.44 & 0.690 & 0.802 & 60.59 & 0.596 & 0.836 \\
NeuroAct & \textbf{85.02} & \textbf{0.822} & \textbf{0.897} & \textbf{61.02} & \textbf{0.601} & \textbf{0.841} \\
\midrule
 & \multicolumn{3}{c}{ADFTD} & \multicolumn{3}{c}{HMC} \\
\cmidrule(lr){2-4}\cmidrule(lr){5-7}
Agent & BA $\uparrow$ & F1 $\uparrow$ & AUC $\uparrow$ & BA $\uparrow$ & F1 $\uparrow$ & AUC $\uparrow$ \\
CodeAct & 58.28 & 0.581 & 0.729 & 74.46 & 0.748 & 0.928 \\
BrainAgent & 52.61 & 0.501 & 0.760 & 74.50 & 0.749 & 0.929 \\
BrainPilot & 62.04 & 0.620 & 0.754 & 73.91 & 0.736 & 0.913 \\
NeuroAct & \textbf{74.89} & \textbf{0.729} & \textbf{0.843} & \textbf{75.14} & \textbf{0.750} & \textbf{0.945} \\
\bottomrule
\end{tabular*}
\end{minipage}\par\medskip

\par\medskip\noindent\begin{minipage}{\linewidth}
\captionsetup{type=table,hypcap=false,skip=5pt}\centering
\caption{Collaborative Prediction HBN results with DeepSeek.}\label{tab:app_collab_agents_hbn}
\footnotesize\setlength{\tabcolsep}{3pt}
\begin{tabular*}{\linewidth}{@{\extracolsep{\fill}}lrrr@{}}
\toprule
Agent & $r\uparrow$ & MAE $\downarrow$ & MSE $\downarrow$ \\
\midrule
CodeAct & 0.760 & 1.649 & 5.106 \\
BrainAgent & 0.758 & 1.661 & 5.158 \\
BrainPilot & 0.767 & 1.661 & 5.144 \\
NeuroAct & \textbf{0.780} & \textbf{1.585} & \textbf{4.702} \\
\bottomrule
\end{tabular*}
\end{minipage}\par\medskip

\FloatBarrier
\section{Controls, Costs, and Transfer}
\label{app:baselines}
\subsection{External framework adaptations}
CodeAct uses the implementation released with BrainBench and its Python execution and feedback loop~\citep{wang2024codeact,zhou2026brainbench}. We use BrainPilot's open-source implementation, in which the Principal Investigator coordinates specialist agents using the framework's tools and research skills~\citep{li2026brainpilot}. BrainAgent uses a paper-method reimplementation of Supervisor-led task decomposition and domain-agent execution~\citep{zhou2026brainagent}, extended with general Python execution for training and prediction programs.

The adapters provide access to task data and model assets, together with interfaces for executing programs and evaluating predictions. Execution logs and metric feedback return to each framework's own research loop, allowing it to revise experiments and select a final predictor.

Setting B requires a shared encoder with task-specific predictors, trained from the supplied LaBraM model and delivered as one checkpoint covering all five targets. BrainAgent uses a custom encoder partly initialized from LaBraM, with per-task sample limits to avoid execution errors during shared-model training. Setting C permits new scientific computing code for preprocessing and model training. Setting D provides member predictions and development labels for learning sample-level prediction selection. External baselines return a member choice per sample, while NeuroAct additionally supports learned convex combinations.

\subsection{Mechanism ablations}
\label{app:v6_selection}
The ablations fix CBraMod and DeepSeek on Workload, BCIC-IV-2a, and ADFTD to examine how individual research capabilities affect optimization within the same backbone family. Each condition removes one capability through runtime permissions or context filtering. Without specialist personas, a single agent performs the research and implementation in one conversation, with access to the remaining research tools. Without EEG diagnostics, the analysis interface provides descriptive data and signal-quality statistics, while task-directed signal and model analyses are unavailable.

The code-editing control allows the agent to select registered training recipes and adjust their exposed parameters; the full framework also permits new scientific computing code in the workspace. Removing runtime skills disables their contextual guidance and retrieval, with tool interfaces and usage descriptions retained. The auxiliary-metric control exposes only the primary BA score, withholding both auxiliary predictive metrics and training dynamics. Full additionally receives macro-F1 and AUROC, together with training-loss and gradient information. Each agent conducts its own experiments in response to the feedback available in its condition.

Tab.~\ref{tab:app_ablations} supplements Tab.~\ref{tab:v6_ablations} with macro-F1 for the same ablation conditions. Each row reports all metrics from the same evaluated predictor.

\begin{table}[!htb]
\captionsetup{skip=5pt}\centering
\caption{Complete predictive metrics of mechanism ablations for the CBraMod with DeepSeek.}\label{tab:app_ablations}
\footnotesize\setlength{\tabcolsep}{3pt}
\begin{tabular*}{\linewidth}{@{\extracolsep{\fill}}lrrrrrrrrr@{}}
\toprule
 & \multicolumn{3}{c}{Workload} & \multicolumn{3}{c}{BCIC-IV-2a} & \multicolumn{3}{c}{ADFTD} \\
\cmidrule(lr){2-4}\cmidrule(lr){5-7}\cmidrule(lr){8-10}
Ablation & BA & F1 & AUC & BA & F1 & AUC & BA & F1 & AUC \\
\midrule
Full NeuroAct & \textbf{77.73} & 0.674 & \textbf{0.886} & \textbf{56.34} & \textbf{0.564} & \textbf{0.805} & \textbf{62.94} & \textbf{0.636} & \textbf{0.831} \\
w/o specialist personas & 77.11 & \textbf{0.694} & 0.853 & 51.65 & 0.504 & 0.778 & 61.29 & 0.600 & 0.723 \\
w/o EEG diagnostics & 72.63 & 0.649 & 0.752 & 52.17 & 0.513 & 0.798 & 41.80 & 0.415 & 0.579 \\
w/o code editing & 66.22 & 0.620 & 0.722 & 52.60 & 0.523 & 0.789 & 48.70 & 0.485 & 0.576 \\
w/o runtime skills & 70.60 & 0.622 & 0.783 & 52.60 & 0.512 & 0.788 & 48.95 & 0.469 & 0.633 \\
w/o auxiliary metrics & 77.29 & 0.673 & 0.862 & 49.74 & 0.475 & 0.786 & 61.62 & 0.609 & 0.829 \\
\bottomrule
\end{tabular*}\end{table}

\noindent\textbf{Research cost.}
Tab.~\ref{tab:app_mechanism_cost} reports the runtime-skill comparisons plotted in Fig.~\ref{fig:v6_efficiency}(a). Each comparison fixes the task, CBraMod, DeepSeek, and resource allowance, and lets the two agents develop their own solutions with or without runtime skills. Both endpoints are the selected predictors from those studies; their costs cover the complete research leading to selection. Tokens sum LLM input and output across research conversations, and GPUh sums accelerator time across experiments.

\begin{table}[!htb]
\captionsetup{skip=5pt}\centering
\caption{Predictive performance and research cost with and without runtime skills. ADFTD time sums active execution intervals.}\label{tab:app_mechanism_cost}
\footnotesize\setlength{\tabcolsep}{3pt}
\begin{tabular*}{\linewidth}{@{\extracolsep{\fill}}llrrrrrr@{}}
\toprule
Dataset & Skills & BA $\uparrow$ & F1 $\uparrow$ & AUC $\uparrow$ & Tokens (M) & GPUh & Min \\
\midrule
Workload & w/o & 70.60 & 0.622 & 0.783 & 25.38 & \textbf{0.159} & 140.3 \\
Workload & with & \textbf{74.68} & \textbf{0.689} & \textbf{0.803} & \textbf{10.10} & 0.174 & \textbf{70.10} \\
PhysioMI & w/o & 51.13 & 0.515 & 0.759 & 14.88 & 0.965 & 173.3 \\
PhysioMI & with & \textbf{62.18} & \textbf{0.621} & \textbf{0.838} & \textbf{9.863} & \textbf{0.687} & \textbf{118.7} \\
ADFTD & w/o & 48.95 & 0.469 & 0.633 & \textbf{21.95} & \textbf{0.320} & 289.1 \\
ADFTD & with & \textbf{57.52} & \textbf{0.568} & \textbf{0.693} & 22.21 & 0.525 & \textbf{276.5} \\
\bottomrule
\end{tabular*}\end{table}

Tab.~\ref{tab:app_bcic_ablations} adds research costs for all BCIC-IV-2a conditions, with research time reported in minutes. The auxiliary-metric controls on Workload and ADFTD consume 24.27M and 25.41M tokens, with 0.315 and 0.611 GPUh, respectively.

\begin{table}[!htb]
\captionsetup{skip=5pt}\centering
\caption{BCIC-IV-2a mechanism ablations and research cost.}\label{tab:app_bcic_ablations}
\footnotesize\setlength{\tabcolsep}{3pt}
\begin{tabular*}{\linewidth}{@{\extracolsep{\fill}}lrrrrrr@{}}
\toprule
Ablation & BA $\uparrow$ & F1 $\uparrow$ & AUC $\uparrow$ & Tokens (M) & GPUh & Min \\
\midrule
Full NeuroAct & \textbf{56.34} & \textbf{0.564} & \textbf{0.805} & 35.11 & \textbf{0.671} & 299.1 \\
w/o specialist personas & 51.65 & 0.504 & 0.778 & \textbf{12.52} & 0.730 & \textbf{127.0} \\
w/o EEG diagnostics & 52.17 & 0.513 & 0.798 & 18.19 & 1.274 & 377.9 \\
w/o code editing & 52.60 & 0.523 & 0.789 & 19.85 & 1.413 & 386.3 \\
w/o runtime skills & 52.60 & 0.512 & 0.788 & 29.62 & 0.767 & 369.6 \\
w/o auxiliary metrics & 49.74 & 0.475 & 0.786 & 29.64 & 0.709 & 368.9 \\
\bottomrule
\end{tabular*}
\end{table}

\subsection{Experience reuse and revision}
\label{app:transfer}
The studies below test how prior research changes a subsequent NeuroAct investigation. We report predictive performance alongside the cost of target research, using DeepSeek throughout.

\textbf{Cross-dataset reuse.} A recommendation from CBraMod research on PhysioMI proposes retaining individual patch representations in the predictor instead of averaging them before classification. We supply this recommendation to a new CBraMod study on BCIC-IV-2a and compare it with research without that experience. The agent investigates its applicability and implements a predictor comparison under the target training recipe. In Tab.~\ref{tab:app_transfer}, experience increases BA from 34.66\% to 50.81\% while reducing research tokens from 29.21M to 10.51M. App.~\ref{app:cases} traces how the recommendation becomes a specific target experiment.

\begin{table}[!htb]
\captionsetup{skip=5pt}\centering
\caption{PhysioMI-to-BCIC-IV-2a experience transfer with CBraMod and DeepSeek.}\label{tab:app_transfer}
\footnotesize\setlength{\tabcolsep}{3pt}
\begin{tabular*}{\linewidth}{@{\extracolsep{\fill}}lllrrrrr@{}}
\toprule
Source & Target & Experience & BA & F1 & AUROC & Tokens (M) & Min \\
\midrule
PhysioMI & BCIC-IV-2a & w/o & 34.66 & 0.278 & 0.592 & 29.21 & 236.7 \\
PhysioMI & BCIC-IV-2a & with & \textbf{50.81} & \textbf{0.495} & \textbf{0.775} & \textbf{10.51} & \textbf{165.3} \\
\bottomrule
\end{tabular*}\end{table}

\textbf{Experience revision.} This comparison tests whether updating a recommendation improves its usefulness in later research. The original advice distinguishes slow convergence from overfitting and proposes a bounded training-schedule experiment for the former. After reviewing additional source curves and model-complementarity analyses, NeuroAct revises the advice to account for slow learning followed by overfitting, reducing the priority of longer training and making prediction combination conditional on suitable validation evidence. Separate CBraMod investigations on PhysioMI receive either the original or revised advice. Tab.~\ref{tab:app_revision} shows that revised advice raises BA from 62.36\% to 63.13\% and reduces research time from 149.8 to 139.1 minutes, with token use increasing from 13.90M to 15.43M.

\begin{table}[!htb]
\captionsetup{skip=5pt}\centering
\caption{Experience revision on PhysioMI with CBraMod and DeepSeek.}\label{tab:app_revision}
\footnotesize\setlength{\tabcolsep}{3pt}
\begin{tabular*}{\linewidth}{@{\extracolsep{\fill}}lrrrrrr@{}}
\toprule
Experience & BA $\uparrow$ & F1 $\uparrow$ & AUC $\uparrow$ & Tokens (M) & GPUh & Min \\
\midrule
Original advice & 62.36 & 0.625 & 0.840 & \textbf{13.90} & 1.096 & 149.8 \\
Revised advice & \textbf{63.13} & \textbf{0.632} & \textbf{0.847} & 15.43 & 0.850 & \textbf{139.1} \\
\bottomrule
\end{tabular*}
\end{table}

\textbf{Cross-model reuse.} These studies keep PhysioMI fixed and transfer experience from CBraMod research to BENDR or BIOT. The recommendation prioritizes checking whether the predictor can use individual patch representations before allocating more computation to training. Each target investigation starts with or without this recommendation and adapts its experiments to the target architecture. For BENDR, Tab.~\ref{tab:app_bendr_transfer} reports the best evaluated predictor from each study and the cost of the complete investigation. Experience reduces tokens from 29.00M to 21.96M, with a BA difference of 0.37 points and an AUROC increase from 0.806 to 0.811. Preparing the source recommendation requires 1.678M tokens.

\begin{table}[!htb]
\captionsetup{skip=5pt}\centering
\caption{BENDR research on PhysioMI with DeepSeek. Scores describe the best evaluated predictors; costs cover each complete target study.}\label{tab:app_bendr_transfer}
\footnotesize\setlength{\tabcolsep}{3pt}
\begin{tabular*}{\linewidth}{@{\extracolsep{\fill}}lrrrrrr@{}}
\toprule
Experience & BA $\uparrow$ & F1 $\uparrow$ & AUC $\uparrow$ & Tokens (M) & GPUh & Min \\
\midrule
w/o & \textbf{58.65} & \textbf{0.587} & 0.806 & 29.00 & 0.950 & 264.1 \\
with & 58.28 & 0.584 & \textbf{0.811} & \textbf{21.96} & \textbf{0.818} & 219.2 \\
\bottomrule
\end{tabular*}
\end{table}

For BIOT, Tab.~\ref{tab:app_biot_transfer} compares the predictors selected by the two investigations. Experience raises BA from 47.92\% to 50.26\% and macro-F1 from 0.459 to 0.504, while reducing tokens from 10.78M to 9.45M and research time from 209.7 to 163.4 minutes. Together, the two model comparisons show that a recommendation can help focus research after changing the encoder, with its benefit reflected in predictive performance or research efficiency.

\begin{table}[!htb]
\captionsetup{skip=5pt}\centering
\caption{CBraMod experience transferred to BIOT on PhysioMI with DeepSeek.}\label{tab:app_biot_transfer}
\footnotesize\setlength{\tabcolsep}{3pt}
\begin{tabular*}{\linewidth}{@{\extracolsep{\fill}}lrrrrrr@{}}
\toprule
Experience & BA $\uparrow$ & F1 $\uparrow$ & AUC $\uparrow$ & Tokens (M) & GPUh & Min \\
\midrule
w/o & 47.92 & 0.459 & 0.730 & 10.78 & 1.679 & 209.7 \\
with & \textbf{50.26} & \textbf{0.504} & \textbf{0.749} & \textbf{9.45} & \textbf{1.243} & \textbf{163.4} \\
\bottomrule
\end{tabular*}
\end{table}

\FloatBarrier
\section{Model Interventions and Continued Research}
\label{app:interventions}
This appendix describes the predictive programs developed by NeuroAct in Sec.~\ref{sec:experiments}. It specifies how the agent implements model and training changes and combines predictions, connecting research proposals to executable interventions.

\subsection{Single-task model development}
The ADFTD CBraMod study tests whether the predictor benefits from access to individual patch representations. The Developer replaces average pooling with a predictor consuming all patch outputs, while retaining the pretrained encoder and full fine-tuning procedure. Both configurations use 20 epochs, batch size 64, and learning rate $5\times10^{-4}$ on the same data split. This revision changes both the information available to the predictor and its capacity, increasing BA from 47.78\% to 60.88\%, macro-F1 from 0.481 to 0.611, and AUROC from 0.570 to 0.752. On BCIC-IV-2a, the selected REVE program similarly uses a flattened predictor, with batch size eight and learning rate $2.4\times10^{-4}$ scaled by 0.1 for the encoder. It reaches 60.59\% BA against the restricted-research reference of 52.60\%.

Training interventions test whether the existing architecture can generalize better under a revised learning procedure. In Workload research, NeuroAct implements Gaussian input perturbations whose magnitude scales with the EEG amplitude; the transformation operates during training, while inference uses the unmodified signal. A separate study responds to majority-class predictions by implementing a loss weighted by training-class frequency. A separate Workload study with REVE explores stronger regularization by increasing weight decay from 0.01 to 0.1 and dropout from 0.3 to 0.5. These interventions change the training program while preserving the prediction task. App.~\ref{app:cases} gives the perturbation formula and the feedback that motivates the loss revision.

Open research also permits the agent to construct a different predictive method. For ADFTD, one program computes relative spectral power and band ratios with regional aggregation, then fits a gradient-boosting classifier. Another implements ShallowFBCSPNet in place of EEGNet while preserving class weights derived from the training data. The latter raises BA from 54.93\% to 71.26\%. These are distinct research solutions, with full results in App.~\ref{app:full_results} and the spectral development trajectory in App.~\ref{app:cases}.

\subsection{Joint training and shared-model refinement}
Shared-model programs jointly train one encoder with task-specific predictors for classification and HBN age regression. Input adapters satisfy each target's input requirements; per-task batch counts, loss weights, and loss scales control contributions to shared updates. Evaluation returns each target's metrics and a joint utility with specified metric directions, scales, and weights. In the LaBraM study, utility is the equal-weight mean of four classification BA fractions and HBN Pearson $r$. After developing a flattened predictor, NeuroAct reduces the encoder learning-rate multiplier from 1 to 0.5 while preserving the remaining configuration, raising validation utility from 0.5184 to 0.5520. App.~\ref{app:full_results} reports initial and revised shared-model results across encoder families.

\subsection{Combination and routing programs}
The programs in Setting D use complementary member predictions at different levels. Workload changes how a pooled probability becomes a class decision, HBN learns a global regression mixture, and ADFTD adapts the mixture to each EEG window.

\noindent\textbf{Workload decision rule.}
Let $p_m(x)$ be member $m$'s positive-class probability and $\mu(x)$ their mean. The training positive-class fraction $q$ determines a threshold $t$, taken as the $\lceil qN\rceil$-th largest pooled score among the $N$ training examples. NeuroAct implements the adjusted score
\begin{equation}
\label{eq:workload_combination}
s(x)=\operatorname{clip}\!\left(\mu(x)+\tfrac12-t,\min_m p_m(x),\max_m p_m(x)\right).
\end{equation}
The shift in Eq.~\ref{eq:workload_combination} moves the decision threshold toward the training-derived operating point. Clipping keeps the score within the range of member probabilities, so sample-dependent convex weights can represent it. The program returns these weights and applies the usual 0.5 decision threshold to the combined probability. Relative to equal weighting, BA increases from 77.95\% to 85.02\%, with AUROC changing from 0.895 to 0.897.

\noindent\textbf{HBN regression mixture.}
NeuroAct fits a nonnegative weight vector summing to one by maximizing Pearson correlation on the development predictions. It then shrinks the vector 10\% toward equal weights. Nested validation grouped by subject compares this mixture with a single-member alternative and determines which to deploy. The resulting vector is fixed across prediction examples, giving $r=0.7801$ and MAE 1.585. The method uses the complementary errors of the members through a shared set of weights.

\noindent\textbf{ADFTD sample-dependent combination.}
The selected program computes 44 features from member posteriors, including confidence and disagreement, and uses a linear softmax gate to assign member weights for each window. The combined class probabilities are the convex sum of the member outputs under these weights. Training minimizes class-weighted cross-entropy with Adam at learning rate 0.05 for 300 steps; subject-grouped validation compares gate configurations before the final fit. The selected gate reaches 74.89\% BA, compared with 71.26\% for the strongest member and 64.44\% for equal weighting. App.~\ref{app:adftd_routing_case} explains the analysis and successive revisions that lead to this program, and App.~\ref{app:full_results} provides complete metrics for all collaborative-prediction tasks.

\FloatBarrier
\section{Case Analyses}
\label{app:cases}
The following trajectories explain how observations changed the research plan and how the resulting implementations were evaluated. Role markers denote the Researcher (R), Developer (D), Reviewer (V), and Coordinator (C).

\subsection{From montage inspection to spectral feature development}
\begin{figure}[!htb]
\centering
\includegraphics[width=\linewidth]{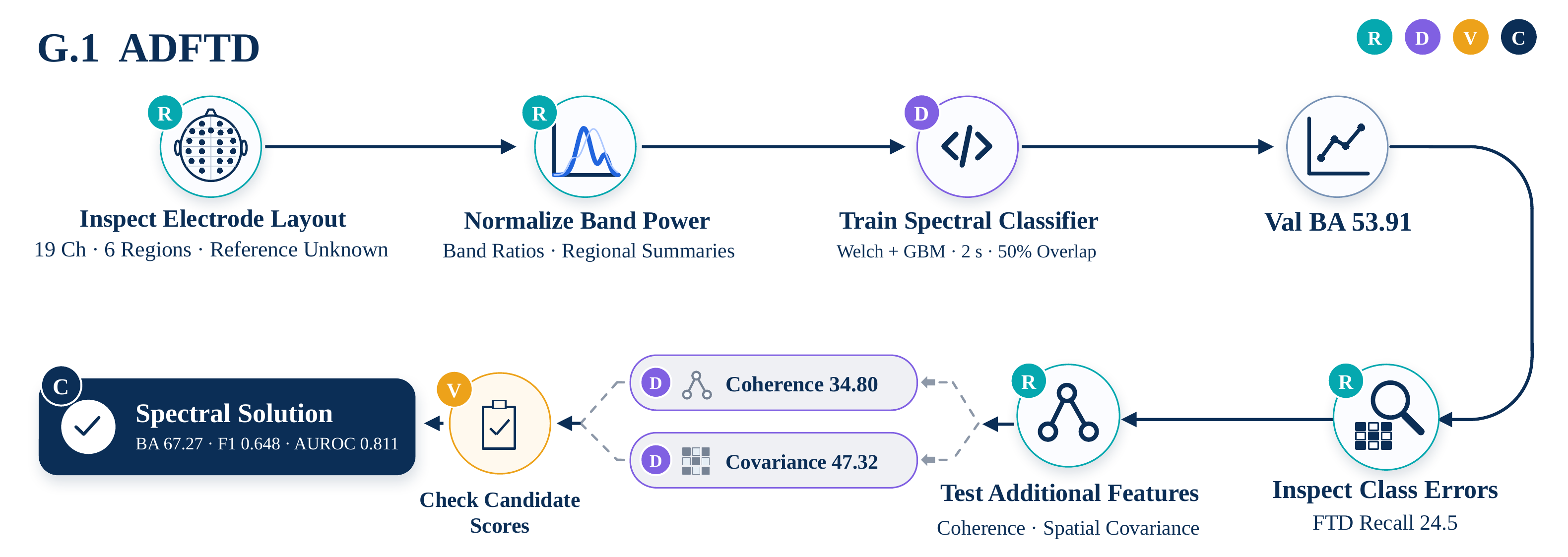}
\captionsetup{skip=4pt}
\caption{ADFTD spectral development guided by montage and prediction-error analysis.}
\label{fig:case_adftd_spectral}
\end{figure}

The ADFTD investigation in Fig.~\ref{fig:case_adftd_spectral} begins by examining the 19-channel montage and the available predictive implementations. The Researcher identifies six scalp regions and notes that the recording reference is unspecified. This favors spectral features normalized within each window, whose construction does not depend on comparing absolute power across recordings. The proposal combines relative band power over 1--45 Hz with logarithmic ratios involving delta, theta, alpha, and beta activity. Regional averages preserve coarse spatial structure, while alpha peak frequency and normalized spectral entropy summarize spectral shape.

The Developer implements a workspace prediction program using Welch spectra with two-second segments and 50\% overlap. Feature standardization is fitted on training data, and a gradient-boosting classifier learns the three-class decision function. The initial validation BA of 53.91\% warrants further investigation. Prediction-error analysis finds that recall for frontotemporal dementia is 24.5\%, motivating a search for information complementary to spectral power. The agent implements coherence features and a separate spatial-covariance predictor; the latter maps shrinkage covariance matrices to a tangent space with a training-fitted reference.

The coherence and covariance alternatives obtain 34.80\% and 47.32\% validation BA, below the spectral predictor's 53.91\%. Their outcomes keep the spectral method as the preferred measured solution, whose test BA is 67.27\% with macro-F1 0.648 and AUROC 0.811. The case demonstrates method construction from EEG conditions and feedback-driven allocation of further experiments, with each tested alternative answering a specific question about the available signal.

\subsection{Input perturbation and generalization}
\begin{figure}[!htb]
\centering
\includegraphics[width=\linewidth]{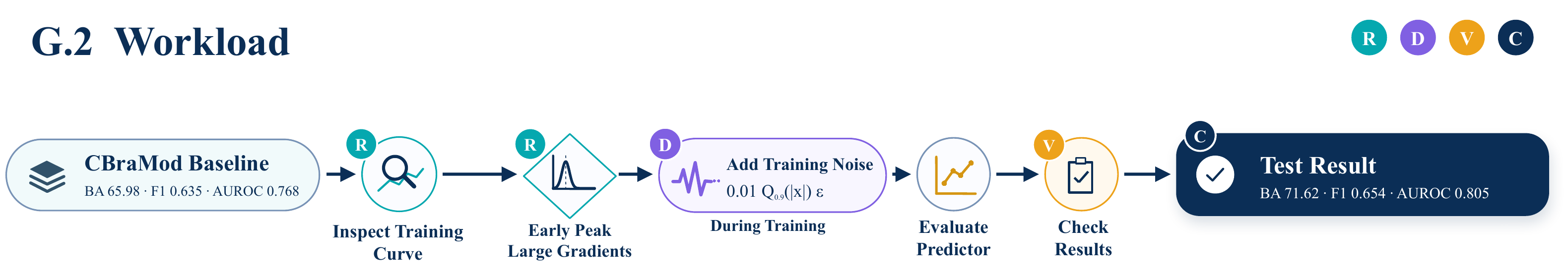}
\captionsetup{skip=4pt}
\caption{Amplitude-scaled training perturbations in the Workload study.}
\label{fig:case_workload_perturbation}
\end{figure}

In the Workload CBraMod study shown in Fig.~\ref{fig:case_workload_perturbation}, the initial learning curve peaks early and then deteriorates, accompanied by large gradient norms and variability across completed training runs. This motivates regularization experiments, including a custom training program that adds Gaussian perturbations scaled to the current EEG tensor. The implementation uses
\begin{equation}
 x'=x+0.01Q_{0.9}(|x|)\epsilon,\qquad \epsilon\sim\mathcal N(0,I),
\end{equation}
where $Q_{0.9}$ is the 90th percentile of absolute input amplitude. Perturbations are applied during training, while prediction uses the unmodified input. This preserves amplitude-relative noise strength as the signal scale changes.

The completed comparison raises test BA from 65.98\% to 71.62\%, macro-F1 from 0.635 to 0.654, and AUROC from 0.768 to 0.805. The case links a diagnosed training behavior to an implemented data-side intervention.

\subsection{Training feedback redirects optimization toward class balance}
\begin{figure}[!htb]
\centering
\includegraphics[width=\linewidth]{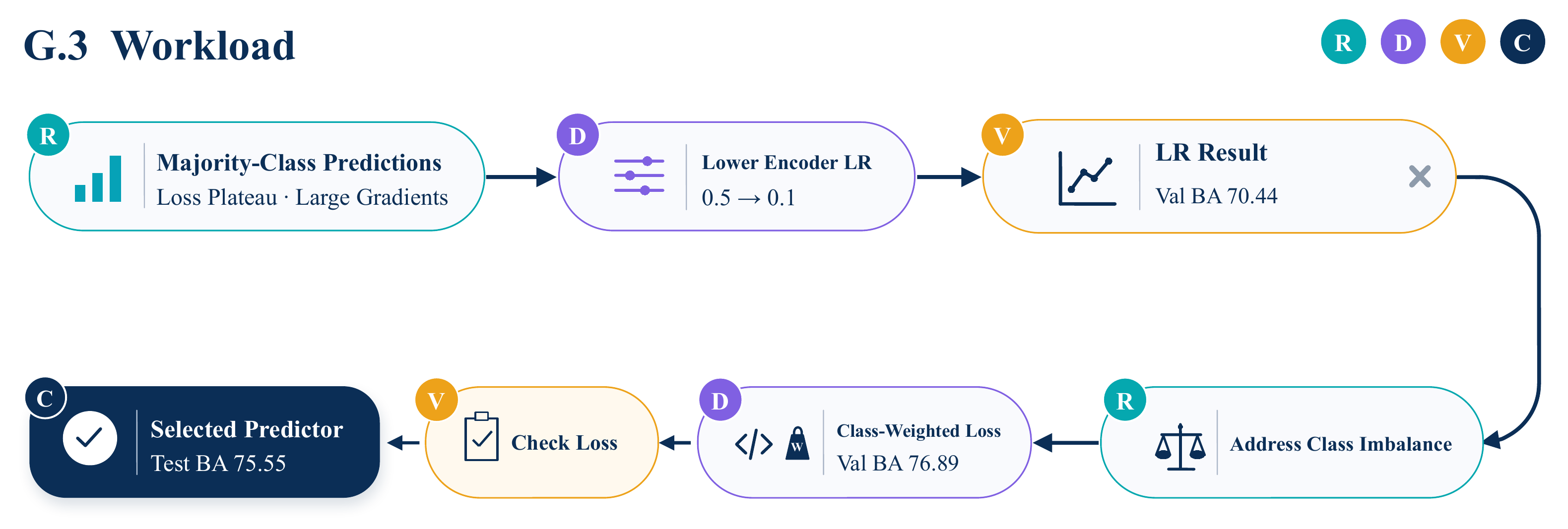}
\captionsetup{skip=4pt}
\caption{Workload research shifts from encoder learning-rate reduction to a class-weighted loss.}
\label{fig:case_workload_loss}
\end{figure}

In the separate Workload investigation summarized in Fig.~\ref{fig:case_workload_loss}, the initial CBraMod predictor concentrates decisions on the majority class as training loss plateaus and gradient norms remain large. The agent first tests whether excessive encoder updates explain this behavior. Reducing the encoder learning-rate scale from 0.5 to 0.1 leaves validation BA at 70.44\%, while the maximum pre-clipping gradient norm rises from 19.12 to 43.15. These observations redirect research toward class imbalance and the loss function.

The Researcher organizes this evidence into a class-balancing proposal, and the Developer implements a loss weighted by training-class frequency. The Reviewer checks the implementation and the comparison before delivery. Validation BA reaches 76.89\%, and the selected predictor attains 75.55\% test BA. The transition from a learning-rate adjustment to an authored loss function illustrates the capability tested by the code-editing ablation.

\subsection{Making transferred experience specific enough to test}
\begin{figure}[!htb]
\centering
\includegraphics[width=0.9\linewidth]{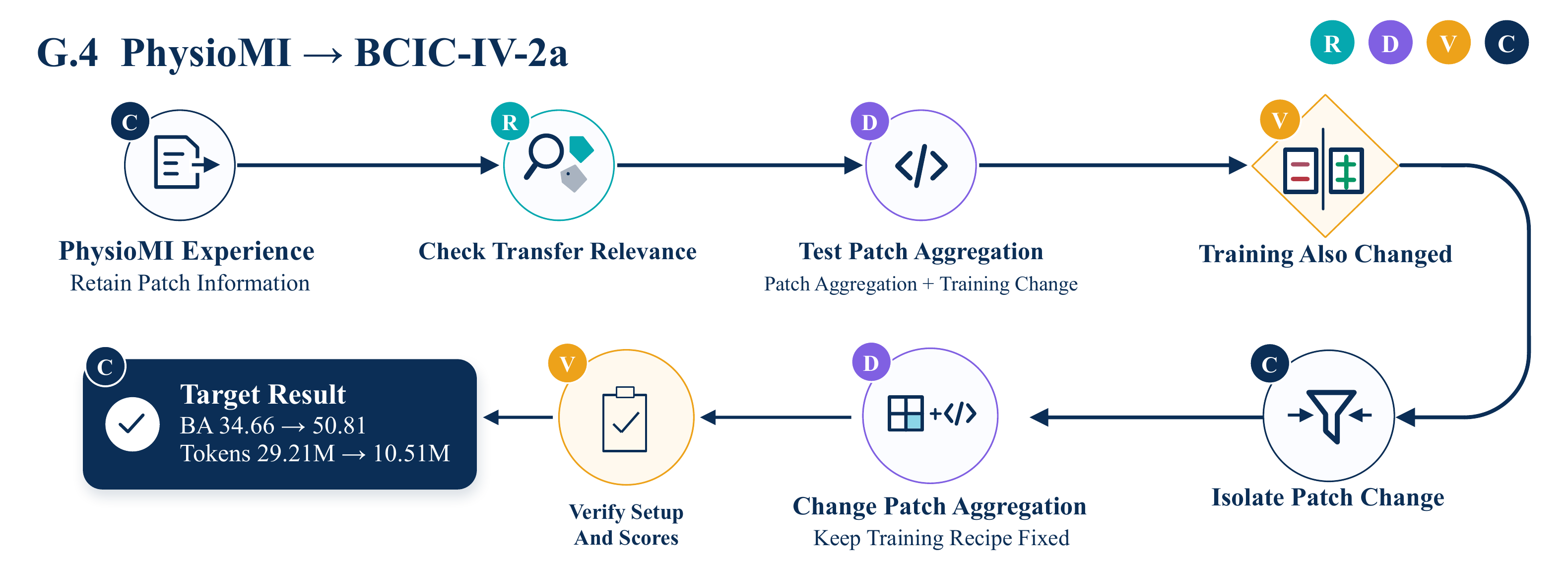}
\captionsetup{skip=4pt}
\caption{Transferred experience guides a patch-aggregation experiment on BCIC-IV-2a.}
\label{fig:case_experience_transfer}
\end{figure}

The source PhysioMI investigation in Fig.~\ref{fig:case_experience_transfer} suggests retaining CBraMod patch information instead of averaging it before prediction. In the BCIC-IV-2a study, this recommendation gives the Researcher a concrete target for investigation. The first implementation also changes the training recipe. To isolate the effect of patch aggregation, the Coordinator narrows the comparison to the predictor while preserving the established recipe.

The revised target implementation reaches 50.81\% BA, compared with 34.66\% without experience, while total research tokens decrease from 29.21M to 10.51M. The trace shows how an experience item guides a new experiment and becomes more precise through target-side checks. The separate revision study in App.~\ref{app:transfer} measures the utility of updating advice before another investigation.

\subsection{Using complementarity at the appropriate prediction stage}
\begin{figure}[!htb]
\centering
\includegraphics[width=\linewidth]{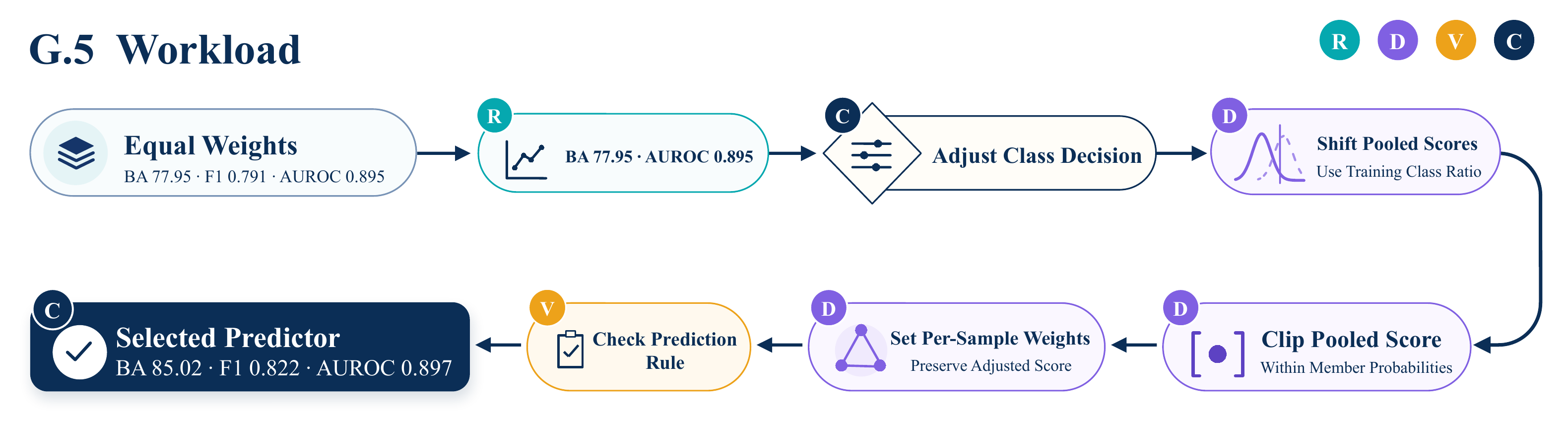}
\captionsetup{skip=4pt}
\caption{Workload collaborative prediction through a revised decision rule.}
\label{fig:case_workload_combination}
\end{figure}

In the Workload collaborative prediction study shown in Fig.~\ref{fig:case_workload_combination}, equal weighting provides an AUROC of 0.895 and 77.95\% BA. The useful intervention therefore concerns how a pooled score produces a class decision. NeuroAct implements a shift based on the training class proportion and clips the adjusted score to the range of member probabilities, allowing sample-dependent convex weights to realize the result. BA reaches 85.02\% and AUROC increases to 0.897, with macro-F1 rising from 0.791 to 0.822.

The HBN study instead fits global convex weights and shrinks them toward equal weighting. Subject-grouped validation compares this mixture with a single-member alternative. The resulting program reaches Pearson $r=0.780$, MAE 1.585, and MSE 4.702. The two studies locate the intervention at different stages of prediction according to the observed model outputs. App.~\ref{app:interventions} details these methods.

\subsection{Turning complementary predictions into sample-dependent decisions}
\label{app:adftd_routing_case}
\begin{figure}[!htb]
\centering
\includegraphics[width=\linewidth]{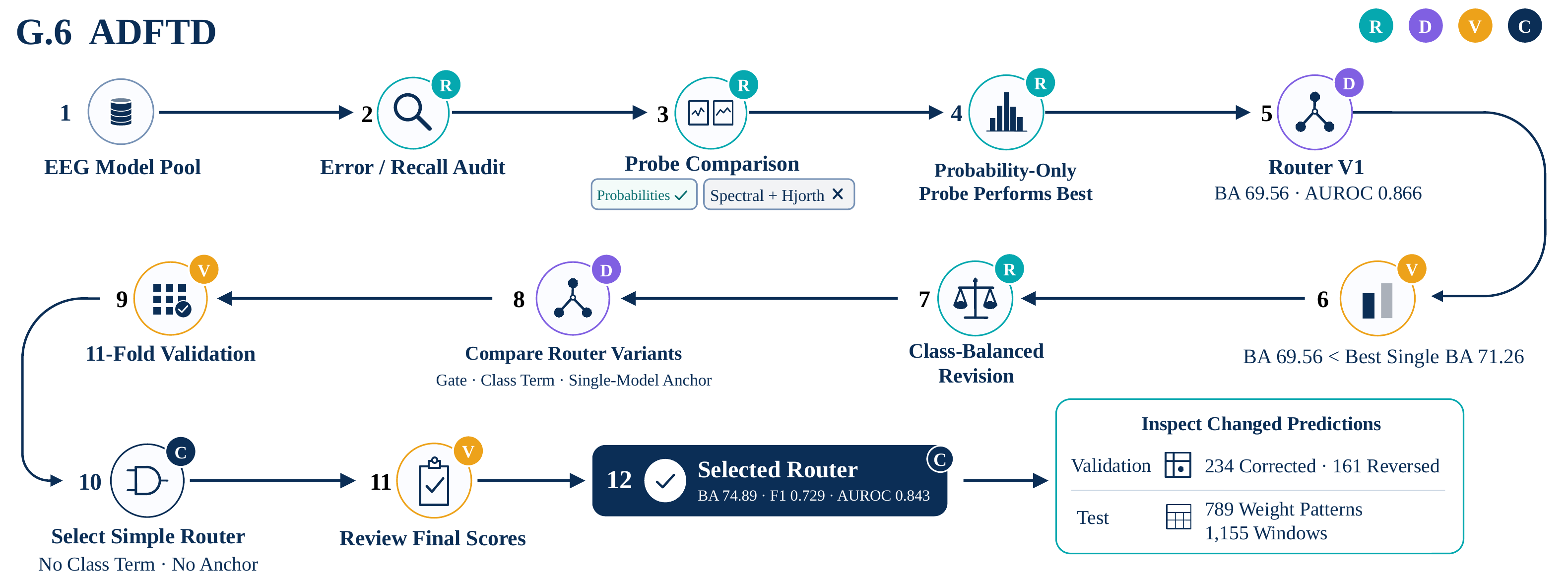}
\captionsetup{skip=4pt}
\caption{ADFTD router development from prediction analysis to a sample-dependent combination.}
\label{fig:case_adftd_routing}
\end{figure}

The ADFTD study in Fig.~\ref{fig:case_adftd_routing} investigates how to combine member predictions into a predictor that improves on the strongest member. The initial investigation compares member errors and class recall, then evaluates routing rules with subject-grouped validation. Posterior-based probes outperform their counterparts with additional spectral and Hjorth features, directing development toward the information already present in model outputs. The first evaluated sample-dependent router achieves 69.56\% BA and 0.866 AUROC. Its strong ranking performance, alongside BA below the strongest member's 71.26\%, leads the Coordinator to request a revision focused on class-balanced decisions.

Using fixed member predictions, the Developer implements convex combinations with per-window softmax weights, an optional class-conditioned component, and shrinkage toward the strongest member. The router uses 44 features summarizing member probabilities, confidence, and disagreement. It fits a class-weighted cross-entropy objective, evaluates held-out subjects in 11 outer folds, and selects configurations using grouped validation within each training fold. The final fit selects the sample-dependent gate without the optional class-conditioned component or anchor shrinkage.

The Coordinator selects the revised router, which reaches 74.89\% BA, 0.729 macro-F1, and 0.843 AUROC. BA improves by 3.63 points over the strongest member and 10.45 points over equal weighting. Frozen deployment weights contain 789 distinct rows at four-decimal precision across 1,155 windows, with every member receiving the largest weight on some windows. An analysis of the outer-fold validation predictions further shows how the decision rule redistributes errors. Relative to ShallowFBCSPNet, it corrects 234 previously incorrect windows and changes 161 correct predictions to errors, raising validation BA from 37.70\% to 45.03\%. Recall for frontotemporal dementia increases from 3.15\% to 20.98\%, alongside changes in the other classes. The case illustrates how analysis of complementary predictions and metric feedback can guide the agent toward a useful sample-dependent combination.

\FloatBarrier
\section{Repeatability of Autonomous Research}
\label{app:stability}
We conducted ten independent NeuroAct studies on each of Workload and BCIC-IV-2a using DeepSeek and CBraMod. Data splits, starting model assets, initial experience, and LLM settings are held fixed, with 50 million tokens and 5 GPU hours available per study. Each study contributes its highest-BA evaluated candidate, with macro-F1 and AUROC taken from that same candidate. Research tokens cover the complete study. Active research time combines recorded LLM, tool, and experiment execution intervals, excluding idle waiting and recovery gaps and counting overlapping activity once. Tab.~\ref{tab:app_stability} summarizes the mean and standard deviation across the ten studies.

\begin{table}[!htb]
\captionsetup{skip=5pt}\centering
\caption{Stability estimates (means $\pm$ standard deviations).}\label{tab:app_stability}
\footnotesize\setlength{\tabcolsep}{3pt}
\begin{tabular*}{\linewidth}{@{\extracolsep{\fill}}lrrrrr@{}}
\toprule
Dataset & BA (\%) $\uparrow$ & Macro-F1 $\uparrow$ & AUROC $\uparrow$ & Tokens (M) $\downarrow$ & Min $\downarrow$ \\
\midrule
Workload & $77.73\pm02.21$ & $0.674\pm0.056$ & $0.886\pm0.017$ & $19.17\pm08.09$ & $106.4\pm25.1$ \\
BCIC-IV-2a & $56.34\pm01.76$ & $0.564\pm0.033$ & $0.805\pm0.021$ & $35.11\pm13.90$ & $209.2\pm54.5$ \\
\bottomrule
\end{tabular*}\end{table}

\end{document}